\documentclass[11pt]{article}
\usepackage[preprint]{acl}
\usepackage{times}
\usepackage{latexsym}
\usepackage[T1]{fontenc}
\usepackage[utf8]{inputenc}
\usepackage{microtype}
\usepackage{inconsolata}
\usepackage{graphicx}
\usepackage{amsmath}
\usepackage{amssymb}
\usepackage{mathtools}
\usepackage{amsthm}

\usepackage{soul}
\usepackage{multirow}
\usepackage{graphicx}
\usepackage{placeins}
\usepackage{booktabs}

\usepackage{tcolorbox}
\tcbuselibrary{most}
\tcbuselibrary{skins}
\definecolor{tcuserbg}{RGB}{245, 245, 245}
\definecolor{tcuserborder}{RGB}{210, 229, 255}
\definecolor{tcuserfont}{RGB}{0, 0, 0}
\tcbset{
  enhanced,
  boxrule=1pt,
  colback=tcuserbg,
  colframe=tcuserborder,
  fonttitle=\bfseries,
  coltitle=tcuserfont,
  boxsep=5pt,
  arc=5pt,
  outer arc=5pt,
  left=10pt,
  right=10pt,
  top=2pt,
  bottom=2pt,
  attach boxed title to top left={yshift=-0.25\baselineskip-1mm,xshift=2mm},
  boxed title style={
    colback=tcuserborder,
    sharp corners,
    rounded corners=northeast,
    arc=3pt,
    outer arc=3pt,
    boxrule=0pt,
    boxsep=2pt,
  },
}

\space{}

\title{\includegraphics[width=0.5cm]{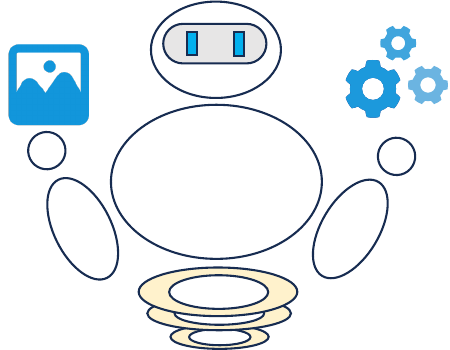}MMGenBench: An Automated Pipeline for Evaluating LMMs from a Text-to-Image Generation Perspective}
\author{Hailang Huang,~
Yong Wang,~
Zixuan Huang,~
Huaqiu Li,\\
\textbf{Tongwen Huang,~
Xiangxiang Chu}\\
\normalfont\href{https://github.com/amap-ml}{AMAP}, Alibaba Group\\
{\small Project Page: \href{https://github.com/lerogo/MMGenBench}{https://github.com/lerogo/MMGenBench}}}

\begin{document}
\clubpenalty=10000
\widowpenalty=10000
\displaywidowpenalty=10000
\maketitle
\begin{abstract}
Existing benchmarks often rely on predefined questions or reference captions, limiting their flexibility for evaluating detailed descriptions on new image collections. We present MMGenBench, an automated benchmark framework for evaluating Large Multimodal Models (LMMs) from raw images without per-image questions or reference captions. The pipeline generates descriptions, reconstructs images with a frozen text-to-image model, and compares the original and reconstructed images using a frozen image encoder. It produces reusable per-image scores and inspectable reconstructions, supporting both model comparison and diagnosis of reconstructible visual information. We construct MMGenBench-Domain with 10,000 generated images and MMGenBench-Test with 1,284 images spanning 13 overlapping visual patterns. Evaluation of 68 LMMs reveals pattern-specific differences in reconstruction quality that overall scores can obscure. On 1,000 human-labeled caption pairs from six models, the default score achieves 90.00\% sample-level agreement. Factual perturbations and component analyses characterize the score's sensitivity under the evaluated pipeline, while a COCO study provides initial validation on natural photographs. MMGenBench complements existing benchmarks by linking automated comparisons to visual evidence of the information preserved in LMM descriptions. Project page: \href{https://github.com/lerogo/MMGenBench}{github.com/lerogo/MMGenBench}.
\end{abstract}\space{}
\begin{figure}[t]
\centering
\includegraphics[width=\linewidth]{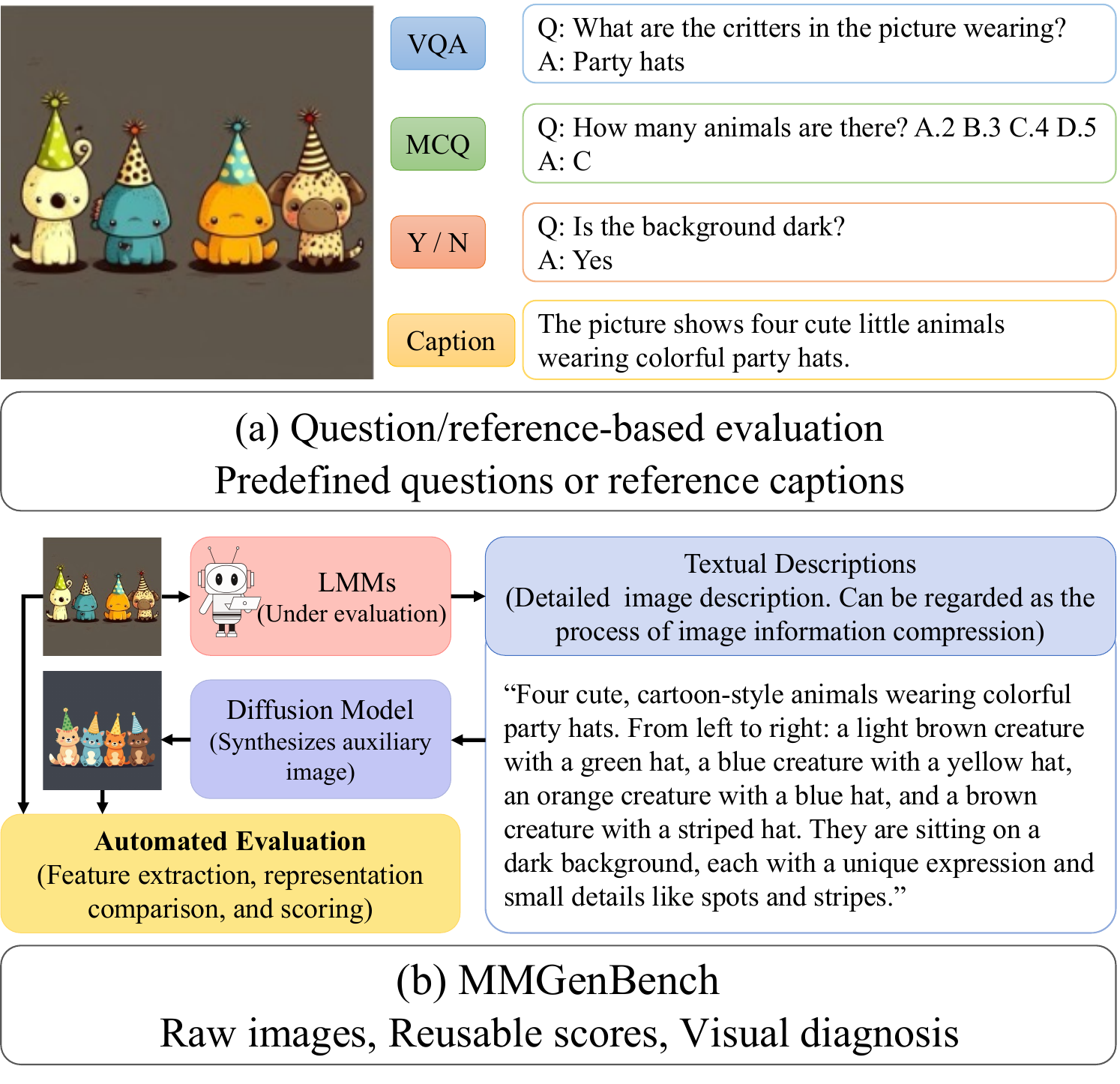}
\caption{Motivation for MMGenBench. Question- or reference-based evaluation uses predefined questions or reference captions. MMGenBench scores detailed descriptions from raw images, producing reusable scores and inspectable reconstructions. The comparison highlights differences in the scoring inputs required by the illustrated evaluation approaches.}
\label{fig:benchmark_comparison}
\end{figure}\space{}
\section{Introduction}
\label{sec:intro}
Large Multimodal Models (LMMs)~\cite{LMM_servey,ovis,Qwen3VL,internvl3_5} extend LLMs~\cite{openai2023gpt,team2023gemini,openai2024gpt4ocard,llama3} to visual-textual processing. The density of visual information and the complexity of spatial relations make it essential to assess how faithfully text preserves image content.

Standard benchmarks~\cite{mmbench,mmstar,mmmu,fu2023mme,blink,OCRBench,megabench,omnidocbench} evaluate tasks including VQA~\cite{MathVista,mmvet,mmlongbench_doc}, captioning~\cite{imagecaptiontask}, and OCR~\cite{OCRBench,ocrv2_bench,ccocr_bench}. Question-based evaluations probe predefined attributes, while reference captions may omit visible details; constructing either for new domains can require substantial annotation. Detailed-caption benchmarks and VLM judges address parts of this problem~\cite{caparena2025}. We ask whether an LMM description preserves enough information to reconstruct its source image. This evaluates the visual information a model chooses to express, beyond what predefined questions elicit.

LMMs convert images into language, while text-to-image models~\cite{flux,sd3,kolors} synthesize visual content from language. A description acts as a selective compression of the image: it encodes selected objects, attributes, and relations in a form that can guide generation. Reconstruction provides a visual probe of the information conveyed by the description, a principle already used by Visual Fact Checker's CLIP-Image-Score~\cite{vfc2024} and CAMScore~\cite{camscore2025}.

We present MMGenBench, which turns this reconstruction principle into a reusable framework for LMM comparison and diagnosis (Fig.~\ref{fig:benchmark_comparison}). MMGenBench-Pipeline prompts LMMs to describe images, reconstructs them with a frozen generator, and compares original and generated images with a pretrained encoder. Its reusable per-description score measures \emph{reconstructible visual information} under fixed components, requiring no per-image questions or reference captions. Inspecting the original image, description, and reconstruction together helps readers locate omitted or altered details and examine whether a discrepancy is already present in the description or appears during reconstruction. This connects model comparisons to concrete output behavior.

We instantiate MMGenBench-Domain with 10,000 JourneyDB images and MMGenBench-Test with 1,284 images covering 13 overlapping visual patterns. Evaluation of 68 LMMs reveals differences across model variants and patterns that overall scores can obscure. Qualitative examples connect these differences to output behavior. A COCO study examines natural photographs; human comparisons and factual, component, and length sensitivity analyses assess the reliability and scope of this diagnostic use.

Our contributions are as follows:
\begin{itemize}
    \item \textbf{Benchmark Framework}: We link reconstruction scores to inspectable outputs in an automated protocol for evaluating LMM descriptions from raw images, without per-image questions or reference captions.
    \item \textbf{Comprehensive Benchmarks}: We construct MMGenBench-Domain for distribution-level evaluation and MMGenBench-Test for diagnosis across 13 visual patterns.
    \item \textbf{Model Analysis and Validation}: We evaluate 68 LMMs, identify differences across model variants and visual patterns, and connect scores to inspectable outputs. Human comparisons and sensitivity studies assess the reliability and scope of these findings.
\end{itemize}\space{}
\begin{figure*}[t]
\centering
\includegraphics[width=\textwidth]{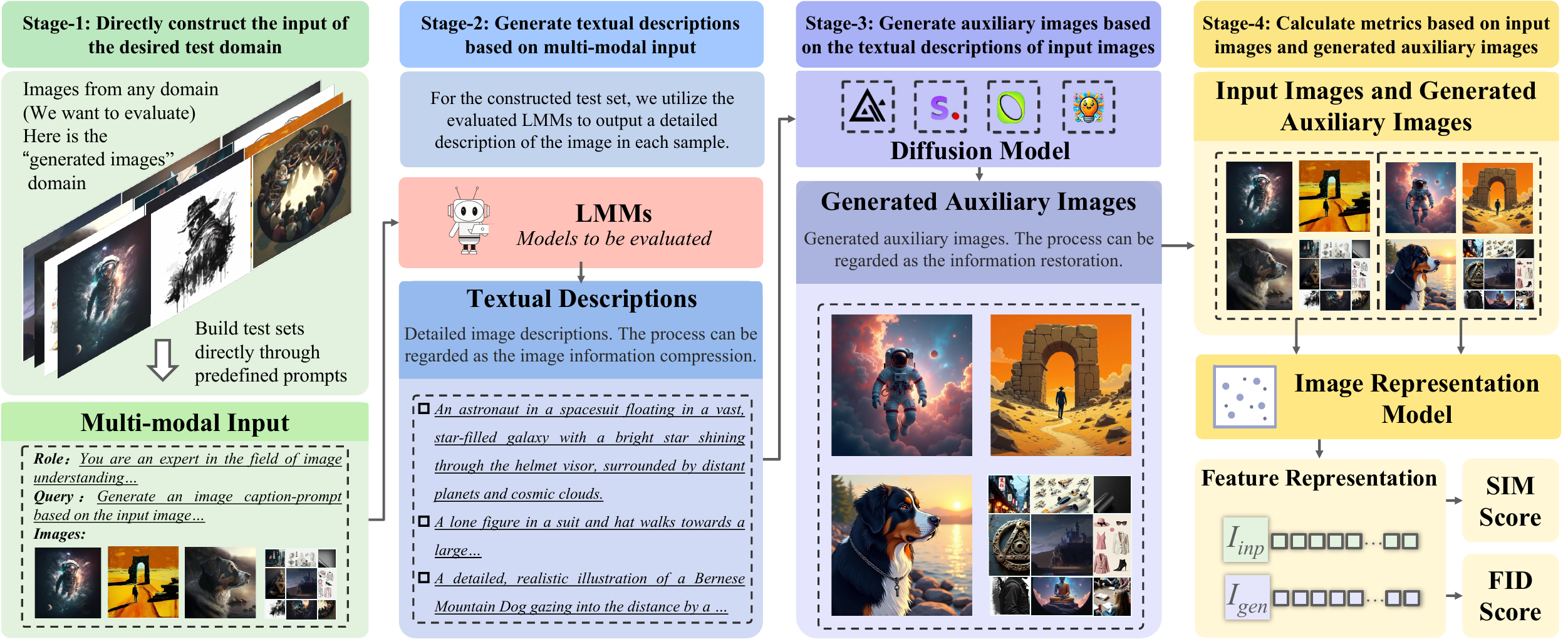}
\caption{MMGenBench-Pipeline. An LMM describes the input image, a frozen text-to-image model produces a reconstruction, and a frozen encoder compares the two images. SIM is the primary metric; the distributional FID-Score is secondary. Scores depend jointly on the description, generator, and encoder.}
\label{fig:pipeline}
\end{figure*}\space{}
\section{Related Work}
\label{sec:related}
\label{appendix:Related-Work}
\noindent\textbf{Automated and Multimodal Benchmarks.}
Automated benchmark construction reduces the cost of evaluating emerging models. AutoBencher~\cite{li2024autobencher} and UNIGEN~\cite{wu2024unigen} generate evaluation data, while Task Me Anything~\cite{taskmeanything} and AUTOBENCH-V~\cite{autobench-v} extend automation to multimodal evaluation. MMStar~\cite{mmstar} and MMMU-Pro~\cite{mmmu_pro} address visual shortcuts and contamination, and BLINK~\cite{blink} and HallusionBench~\cite{hallusion_bench} diagnose perceptual errors. These benchmarks provide complementary task-specific evidence. MMGenBench instead examines information preserved in open-ended descriptions and requires only images at scoring time; reusable scores and reconstructions connect model comparisons to \mbox{visual diagnosis}.

\noindent\textbf{Caption Evaluation and Reconstruction.}
Visual Fact Checker~\cite{vfc2024} introduces CLIP-Image-Score, which generates an image from a caption and compares its CLIP embedding with that of the original. Our CLIP-encoder variant instantiates this scoring principle on the same reconstructed images. D2I-Bench in Image Textualization~\cite{pi2024image} likewise evaluates detailed descriptions through text-to-image reconstruction and image-feature similarity. Building on reconstruction-based caption evaluation, MMGenBench combines 68-model coverage, a 13-pattern resource, and diagnostic validation in a shared evaluation framework. CAMScore~\cite{camscore2025} compares cyclic reconstructions at multiple visual levels. ITIScore~\cite{itiscore2026} combines the original image, reconstruction, and caption through an MLLM and a scoring head trained using human mean opinion scores. MMGenBench uses frozen components without metric-specific training or caption-level supervision. CapArena~\cite{caparena2025} studies detailed captions through human pairwise preferences and introduces CapArena-Auto using VLM judging. SIM assigns each caption an independent scalar score under a fixed configuration, supporting reuse across pairwise comparisons and pattern-level analyses. We compare with a reconstruction baseline and zero-shot VLM judges.\space{}
\section{The MMGenBench-Pipeline}
\label{sec:benchmark}
\subsection{Automatic Evaluation Pipeline}
The pipeline comprises four sequential stages: input construction, description generation, image reconstruction, and scoring (Fig.~\ref{fig:pipeline}). Frozen text-to-image and image representation models provide a shared configuration for comparing LMMs.

\noindent\textbf{Input and Description Generation.}
Given an image from the target domain, we construct the multimodal input using a fixed prompt $\mathbf{P}_{art}$ with five components: role, definition, task, key points, and output format (Appendix~\ref{appendix:Evaluation-Pipeline-Prompt}). The prompt requests a description of 20--60 words suitable for text-to-image generation. The LMM produces
\begin{equation}
    \mathbf{P}_{gen}=\operatorname{LMM}(\mathbf{I}_{inp},\mathbf{P}_{art}).
    \label{equ:1}
\end{equation}
No reference caption or per-image question is required. The requested generator-compatible wording also makes instruction following and description style relevant to the score.

\noindent\textbf{Auxiliary Image Generation.}
A frozen text-to-image model reconstructs an auxiliary image:
\begin{equation}
    \mathbf{I}_{gen}=G(\epsilon;\mathbf{P}_{gen},\theta),
    \label{equ:2}
\end{equation}
where $\epsilon$ is the sampled latent variable and $\theta$ denotes the generator parameters. FLUX.1-dev~\cite{flux} is our default. Stable Diffusion~3.5~\cite{sd3}, Kolors~\cite{kolors}, and Lumina~\cite{lumina} are evaluated separately in the component sensitivity analysis.

\noindent\textbf{Image Representation.}
Generation stochasticity makes pixel-level matching unsuitable for this setting. We therefore extract representations using a frozen encoder $E$:
\begin{equation}
\label{equ:3}
    \mathbf{f}_{inp}=E(\mathbf{I}_{inp}),\quad
    \mathbf{f}_{gen}=E(\mathbf{I}_{gen}).
\end{equation}
We use Unicom~\cite{unicom} by default because its instance-level retrieval objective is suited to comparing an original image with its reconstruction. We also evaluate CLIP~\cite{clip} and DINOv2~\cite{dinov2} (\mbox{Appendix~\ref{sec:Sensitivity-Analysis-of-Embedding-Models}}).

\subsection{Evaluation Metrics and Interpretation}
\label{sec:metrics}
\noindent\textbf{Primary Metric: SIM-Score.}
We compute cosine similarity between the image features:
\begin{equation}
\label{equ:sim}
    \operatorname{SIM}(\mathbf{I}_{inp},\mathbf{I}_{gen})=
    \frac{\mathbf{f}_{inp}\cdot\mathbf{f}_{gen}}
    {\|\mathbf{f}_{inp}\|\|\mathbf{f}_{gen}\|}.
\end{equation}
SIM ranges from $-1$ to $1$ and measures alignment in the selected representation space; it is not a calibrated probability of caption correctness. We average per-image SIM for benchmark and pattern scores. With CLIP as the encoder, this follows the reconstruction scoring principle of Visual Fact Checker~\cite{vfc2024}. With any encoder, a low score can arise from an incomplete description, a rendering failure, or encoder insensitivity. We accordingly interpret SIM as preservation of reconstructible visual information under a fixed configuration, with contributions from description, rendering, and representation.

Each score links an image, description, and reconstruction for visual inspection. Benchmark averages summarize per-image outcomes; pattern labels organize them by content. Human comparisons assess complete descriptions, controlled factual variants test sensitivity to specified changes, and component sweeps examine generator and \mbox{encoder effects}.

\noindent\textbf{Secondary Distributional Metric.}
We additionally compute a Fr\'echet distance between the original and generated feature distributions:
\begin{equation}
\label{equ:fid}
\begin{split}
    \operatorname{FID\text{-}Score}={}&\|\mu_x-\mu_y\|^2\\
    &+\operatorname{Tr}\bigl(\Sigma_x+\Sigma_y
      -2(\Sigma_x\Sigma_y)^{1/2}\bigr),
\end{split}
\end{equation}
where $\mu$ and $\Sigma$ denote feature means and covariances. FID-Score is a Fr\'echet distance in encoder space, using Unicom by default. It differs from standard Inception FID, and values across encoders are not directly comparable. We retain it mainly for the 10,000-image MMGenBench-Domain. Test and per-pattern estimates use smaller samples and are reported only as exploratory results in the appendix. All primary comparisons use SIM.

\section{The MMGenBench Benchmark}
\label{sec:MMGenBench-Benchmark-Construction}
MMGenBench-Pipeline scores new image collections without per-image questions or reference captions, although metric validity still requires domain-specific validation. We construct \textbf{MMGenBench-Domain} by randomly sampling 10,000 images from the JourneyDB~\cite{sun2024journeydb} validation set. It evaluates descriptions of generated images and supports distributional analysis. Section~\ref{sec:Real-World-Domain-Image-Evaluation} additionally examines a natural-photo subset from COCO~\cite{microsoftcoco}.

\begin{figure}[ht]
\centering
\includegraphics[width=.90\linewidth]{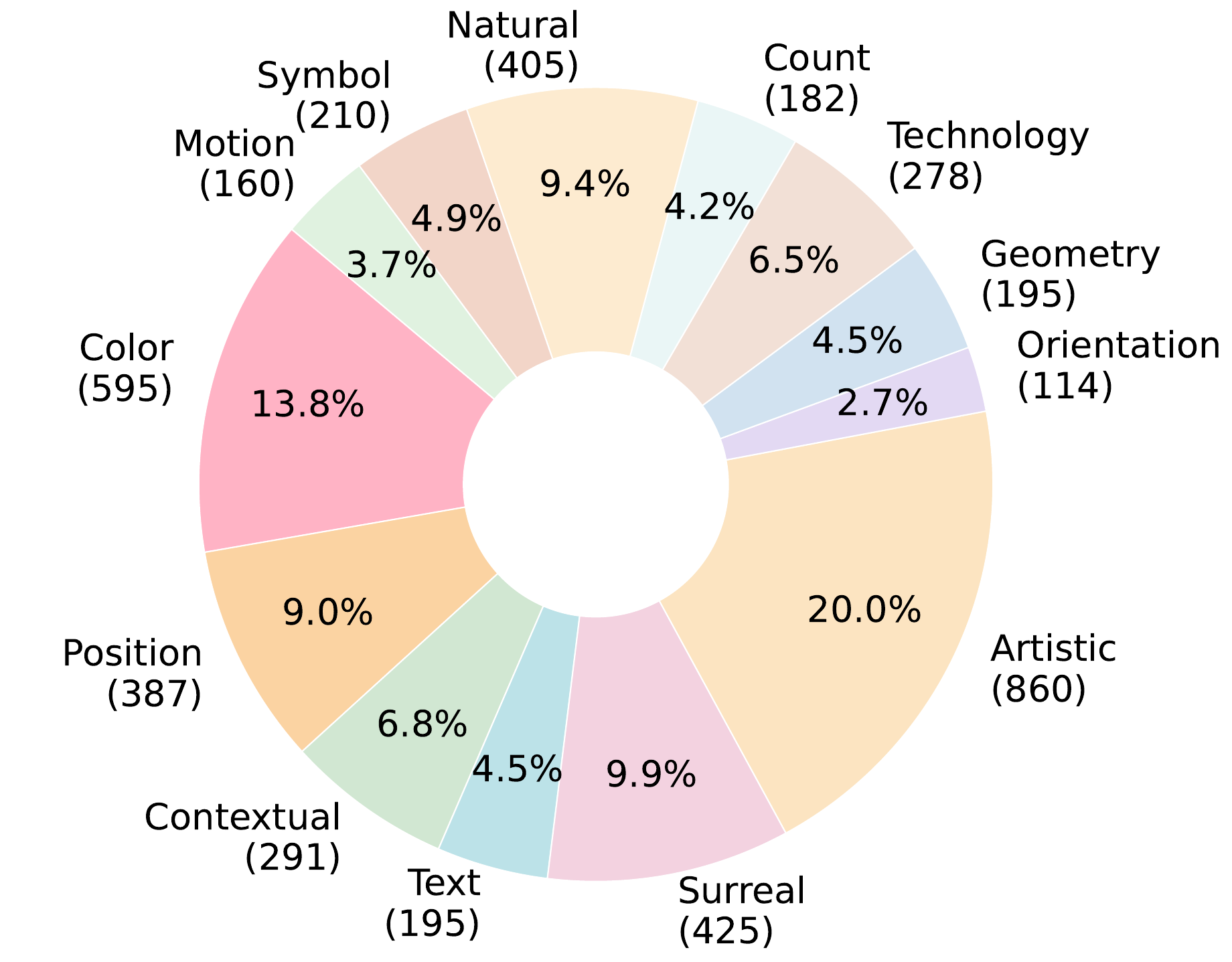}

\caption{Statistics of MMGenBench-Test, which contains $1{,}284$ images across $13$ image patterns, with per-pattern sample counts shown. Percentages refer to pattern assignments, since labels overlap. See Sec.~\ref{sec:MMGenBench-Benchmark-Construction} for construction details.}
\label{fig:sample_statistics}
\end{figure}

For finer-grained diagnosis, \textbf{MMGenBench-Test} contains 1,284 JourneyDB test images spanning 13 overlapping visual patterns. GPT-4o first extracts candidate patterns, GPT-4-Turbo consolidates them into a taxonomy, and GPT-4o re-annotates images under this fixed schema. The final labels incorporate manual checks. Each pattern contains at least 114 images, and an image may have multiple labels (Fig.~\ref{fig:sample_statistics}). These labels organize per-image scores into interpretable profiles of the visual content preserved by each model. Full construction details and prompts appear in Appendix~\ref{appendix:MMGenBench-Benchmark-Details}.
Domain supports distributional analysis, while Test supports pattern-level diagnosis. Pattern scores average SIM over images carrying each label. Because labels overlap, their counts need not sum to 1,284, and averaging pattern means does not recover the overall Test score, which weights each image equally regardless of its labels.
\space{}\space{}
\section{Experiment}
\label{sec:Experiment}
\subsection{Experimental Setup}
\label{sec:Experimental-Setup}
\noindent\textbf{Evaluation Models.}
We evaluate 68 LMMs, including Qwen2-VL~\cite{Qwen2VL}, Qwen3-VL~\cite{Qwen3VL}, InternVL2~\cite{internvl}, InternVL3.5~\cite{internvl3_5}, LLaVA-OV~\cite{llavaonevision}, Ovis~\cite{ovis}, and closed-source API models. Full model coverage appears in Tables~\ref{tab:suppl-results-all} and~\ref{tab:suppl-results-test}. The default pipeline uses FLUX.1-dev and Unicom; sensitivity analyses cover four generators and three encoders.

\noindent\textbf{Implementation Details.}
We use the model configurations in VLMEvalKit~\cite{duan2024vlmevalkit}, replacing the query with our fixed prompt. The prompt requests 20--60 words, but actual outputs can depart from this range; length effects are examined in Sec.~\ref{sec:length-bias}. Eight NVIDIA H20 GPUs provide data parallelism during auxiliary image generation and feature extraction. Each reconstruction runs on one GPU. Appendix~\ref{appendix:Experimental-Setup} gives further setup details, and Sec.~\ref{sec:cost} reports measured computation excluding LMM caption generation.

The records include all 87,312 Test descriptions, per-image SIM scores, and source-image IDs. A shared image index preserves these pairings and links each description and score to its final pattern labels for per-image analysis.

\subsection{Model Comparison and Diagnosis}
\label{sec:Main-Results}
\label{sec:Analysis}
\begin{table}[t]
\centering
\small
\setlength{\tabcolsep}{3pt}
\begin{tabular}{lrrrr}
\toprule
\multirow{2}{*}{Model} & \multicolumn{2}{c}{Test} & \multicolumn{2}{c}{Domain} \\
 & SIM$\uparrow$ & Words & SIM$\uparrow$ & FID$\downarrow$ \\
\midrule
\multicolumn{5}{l}{\textit{Closed-source API models}} \\
\textbf{Claude Opus 4.6} & 0.663 & 49.2 & - & - \\
\textbf{Qwen3-VL-Plus} & 0.657 & 55.9 & - & - \\
Qwen3-VL-Flash & 0.651 & 58.5 & - & - \\
\textbf{GPT-5.4} & 0.637 & 46.1 & - & - \\
GPT-5.2 & 0.629 & 48.2 & - & - \\
Qwen2.5-VL-Max & 0.604 & 38.4 & - & - \\
Qwen2.5-VL-Plus & 0.588 & 42.5 & - & - \\
GPT-4o & 0.567 & 36.7 & - & - \\
Qwen-VL-Max & 0.553 & 33.0 & - & - \\
Qwen-VL-Plus & 0.475 & 33.1 & - & - \\
\midrule
\multicolumn{5}{l}{\textit{Open-source models}} \\
\textbf{Qwen3-VL-32B} & 0.637 & 56.0 & 0.635 & 0.573 \\
Qwen3-VL-8B & 0.610 & 40.9 & 0.608 & 0.617 \\
Qwen3-VL-30B-A3B & 0.608 & 38.1 & 0.607 & 0.616 \\
\textbf{InternVL2-76B} & 0.600 & 57.6 & 0.599 & 0.632 \\
InternVL3.5-38B & 0.584 & 48.4 & 0.586 & 0.652 \\
\textbf{Ovis1.6-9B} & 0.583 & 55.6 & 0.579 & 0.667 \\
Qwen2-VL-72B & 0.553 & 33.7 & 0.545 & 0.710 \\
\textbf{MiniCPM-V2.6-8B} & 0.549 & 33.0 & 0.545 & 0.710 \\
Ovis1.5-9B & 0.522 & 78.7 & 0.524 & 0.808 \\
\textbf{LLaVA-OV-SI-72B} & 0.513 & 22.1 & 0.514 & 0.813 \\
LLaVA-OV-72B & 0.494 & 17.9 & 0.491 & 0.872 \\
\textbf{xGen-MM-4.4B} & 0.415 & 95.1 & 0.412 & 0.940 \\
\bottomrule
\end{tabular}
\caption{Representative results ordered by Test SIM within each category. Bold names indicate the best listed variant within a family (singletons are also bold). Words denotes mean caption length. Ranks are point estimates. FID is secondary on Domain; exploratory Test FID and all 68 models appear in Table~\ref{tab:suppl-results-all}.}
\label{tab:main-results-all}
\end{table}\space{}
\noindent\textbf{Overall Performance.}
Table~\ref{tab:main-results-all} compares representative LMMs on Test and Domain, using SIM as the primary metric. Claude Opus 4.6 has the highest Test point estimate ($0.663$), followed by Qwen3-VL-Plus ($0.657$) and Qwen3-VL-Flash ($0.651$). Qwen3-VL-32B reaches $0.637$, matching the displayed score of GPT-5.4. This shows that competitive reconstruction quality is attainable with an open-source model. Section~\ref{sec:Sensitivity-Analysis} examines uncertainty in close comparisons.

\noindent\textbf{Differences across Model Variants.}
Later variants in several families obtain higher SIM: Qwen3-VL-32B improves over Qwen2-VL-72B, and GPT-5.4 improves over GPT-4o. The Ovis variants illustrate why parameter count alone is insufficient: Ovis1.6-9B scores higher than Ovis1.5-9B despite producing shorter descriptions on average. LLaVA-OV-SI also outperforms LLaVA-OV at the listed 72B scale. Figure~\ref{fig:case-model-cmp} further shows that larger Qwen2-VL variants obtain higher scores, whereas InternVL2 does not improve monotonically at every scale. These comparisons distinguish reconstruction quality across model generations and among similarly sized variants. Inspecting descriptions and reconstructions alongside parameter counts connects the observed differences to output behavior under the common pipeline.

\begin{figure}[t]
\centering

\includegraphics[width=0.96\linewidth]{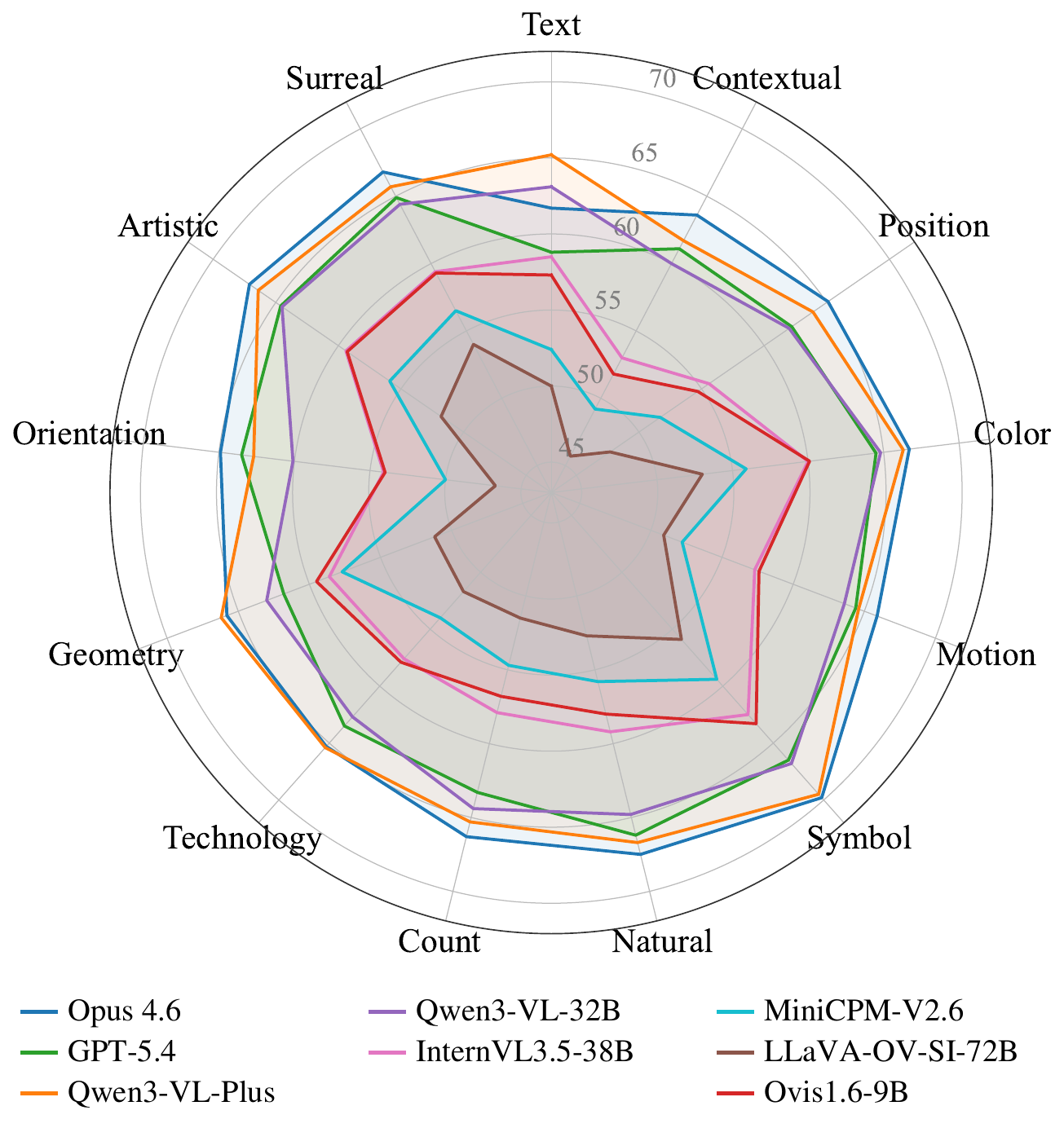}
\caption{Model performance for representative evaluated models across image patterns on MMGenBench-Test. The radial axis is $100\times$SIM. Table~\ref{tab:pattern-selected} gives full model names and reported unscaled values.}
\label{fig:image-pattern-result-cmp}
\end{figure}

\noindent\textbf{Pattern-Level Diagnosis.}
Fig.~\ref{fig:image-pattern-result-cmp} compares representative models across the 13 patterns; Table~\ref{tab:pattern-selected} provides numerical values. ``Symbol'', ``Surreal'', and ``Artistic'' often obtain higher SIM than ``Contextual'' and ``Orientation''. ``Text'' is relatively weak for Claude Opus 4.6 and GPT-5.4: their scores are $0.617$ and $0.588$, compared with $0.698$ and $0.665$ on ``Symbol''. These differences identify content that is less successfully preserved through the complete pipeline, with contributions from the description, generator, and encoder.
The pattern profiles also reveal differences that an overall mean can obscure. Qwen3-VL-Plus obtains $0.652$ on Text, compared with $0.617$ for Claude Opus 4.6, although Claude has the higher overall mean. On Contextual, their scores are $0.617$ and $0.636$, respectively. Likewise, Qwen3-VL-32B and GPT-5.4 have nearly identical overall means, but their Text scores are $0.631$ and $0.588$. These descriptive profiles locate differences within overlapping image subsets and guide inspection of descriptions and reconstructions under the common pipeline.
\space{}

\begingroup
\clubpenalty=10000
\noindent\textbf{Qualitative Diagnosis.}
The examples in Fig.~\ref{fig:analysis-case} and Appendix~\ref{appendix:Qualitative-Example} connect numerical comparisons to observable output behavior. They illustrate three ways to inspect model behavior.
\par\endgroup

First, format deviations remain visible: RBDash prepends an introductory phrase, while MiniCPM adds explanatory prose after the requested description. These outputs show why instruction following matters when a description also serves as a \mbox{generation prompt}.

Second, descriptions differ in the visual details they preserve. LLaVA-OV-72B gives a short description of the puppet scene. In the aircraft example, both Qwen2-VL variants mention the canyon, but the larger model additionally specifies aircraft colors and cockpit details. Inspecting the paired reconstructions lets readers examine which stated details appear in the generated image, beyond the scalar score.

Third, some outputs contain refusals. The xGen-MM animal-scene response begins with a refusal but still includes descriptive content. Reading the response alongside its reconstruction distinguishes refusal behavior from the visual details expressed in the description. Together, these cases show how scores, pattern profiles, and visual inspection complement one another when comparing \mbox{LMM descriptions}.

\noindent\textbf{An Empirical Performance Anchor.}
Let $s_{im}$ denote SIM for image $i$ and model $m$. Selecting the highest-scoring description independently for each image gives the empirical envelope
\begin{equation}
\label{equ:envelope}
    S_{\mathrm{env}}=\frac{1}{N}\sum_{i=1}^{N}\max_{m\in\{1,\ldots,68\}}s_{im},
\end{equation}
where $N=1,284$. The envelope reaches $0.768$, compared with $0.663$ for the best single model. This $0.105$ gap provides an empirical anchor within the current candidate pool. Since selection uses SIM itself, the result is an optimistic envelope rather than a human-caption oracle or theoretical pipeline ceiling. Appendix~\ref{appendix:empirical-envelope} discusses its scope and the available source prompts.

\subsection{Evaluation on Natural Photographs}
\label{sec:Real-World-Domain-Image-Evaluation}
We additionally evaluate four LMMs on 2,000 natural photographs from COCO~\cite{microsoftcoco} (Table~\ref{domain_image_evaluation:1}). In a separate human check, 1,000 comparisons receive three annotations each, with 2,763/3,000 matching votes (92.1\% vote-level agreement). This offers initial evidence for natural photographs beyond JourneyDB's generated-image distribution, within the four-model study.
\begin{table}[t]
\centering
\small
\begin{tabular}{lrr}
\toprule
Model & Test SIM$\uparrow$ & COCO SIM$\uparrow$ \\
\midrule
InternVL2-8B & 0.5478 & 0.4925 \\
Ovis1.6-Gemma2-9B & 0.5825 & 0.5441 \\
Qwen2-VL-7B & 0.5323 & 0.4984 \\
LLaVA-OV-7B & 0.4888 & 0.4344 \\
\bottomrule
\end{tabular}
\caption{Four-model validation on 2,000 COCO photographs. Scores summarize reconstruction quality within each dataset; differing image distributions preclude a direct comparison of task difficulty.}
\label{domain_image_evaluation:1}
\end{table}\space{}

\subsection{Human Preference Alignment and Baseline Evaluation}
\label{sec:Comparison-with-CLIP}
\label{sec:human}
We assess human alignment on 1,000 caption pairs from six representative LMMs, covering all 15 unordered model pairs and 702 distinct images. Nine trained annotators independently compare each pair against the source image; at least five votes define a majority. The one pair without a majority counts as a non-agreement. The SIM tie threshold, $|\Delta|\leq0.01$, was fixed before inspecting human labels. Appendix~\ref{appendix:Human-Alignment-Score} gives the model list, recruitment, training, and aggregation details.

MMGenBench agrees with at least five annotators on 900/1,000 pairs (90.00\%); 7,944/9,000 individual votes agree (88.27\%). Fleiss' $\kappa$ and nominal Krippendorff's $\alpha$ are both $0.940$. These descriptive rates support alignment on this six-model set; repeated images and votes introduce dependence.

\begin{table}[t]
\centering
\small
\setlength{\tabcolsep}{3pt}
\begin{tabular}{lrrr}
\toprule
Method & Sample & Vote & Order \\
\midrule
Direct CLIP image-caption & 60.70 & 61.06 & -- \\
VFC-style CLIP reconstruction & 87.60 & 85.82 & -- \\
MMGenBench (Unicom) & \textbf{90.00} & \textbf{88.27} & -- \\
GPT-5.5 judge & 82.05 & 80.61 & 93.30 \\
Claude Opus 4.8 judge & 80.00 & 78.52 & 90.50 \\
\bottomrule
\end{tabular}
\caption{Human agreement (\%) on the same 1,000 caption pairs. Sample: at least five agreeing annotators; Vote: agreement with individual votes; Order: consistent decisions after swapping captions. Judge agreement is averaged over both orders.}
\label{tab:comparison-with-clip}
\end{table}\space{}
We compare direct CLIP image-caption similarity, a VFC-style CLIP reconstruction score, and two zero-shot VLM judges on the same 1,000 pairs (Table~\ref{tab:comparison-with-clip}). The reconstruction baseline uses CLIP embeddings of the same original and reconstructed images. Its reported agreement is substantially higher than direct CLIP, while Unicom is 2.40 percentage points higher than CLIP reconstruction (90.00\% versus 87.60\%). These results favor reconstruction-based scoring under the tested protocols; the comparison does not isolate the encoder's contribution.

For GPT-5.5 and Claude Opus 4.8 judging, we evaluate both caption orders, map reversed predictions back to caption identities, and average agreement across the two runs. Decisions change on 67 and 95 pairs, respectively. MMGenBench obtains higher agreement under this fixed zero-shot comparison protocol. SIM assigns an independent score to each caption, supporting reuse across different pairings without a presentation-order dependence.

\noindent\textbf{Agreement by Score Margin.}
Table~\ref{tab:human-margin} shows how agreement varies with the per-example absolute margin $|\Delta|$. The 665 comparisons with margins above $0.05$ reach 94.29\% sample agreement, while the 76 comparisons within the tie band reach 72.37\%. The two intermediate groups fall between these values. Thus, the aggregate agreement reflects both clearly separated descriptions and more difficult, similar-score comparisons.
\begin{table}[t]
\centering
\small
\begin{tabular}{lrrr}
\toprule
Absolute SIM margin & Pairs & Sample (\%) & Vote (\%) \\
\midrule
$[0,0.01]$ & 76 & 72.37 & 71.64 \\
$(0.01,0.03]$ & 142 & 83.80 & 83.65 \\
$(0.03,0.05]$ & 117 & 84.62 & 84.05 \\
$>0.05$ & 665 & 94.29 & 91.90 \\
\bottomrule
\end{tabular}
\caption{Human agreement by per-example absolute SIM margin. Counts sum to the same 1,000 comparisons used for the aggregate analysis.}
\label{tab:human-margin}
\end{table}\space{}
These strata reuse the fixed threshold and human labels; their rates are descriptive, not calibrated confidence. Per-example margins concern individual caption comparisons, whereas the paired bootstrap below examines uncertainty in benchmark means.

\subsection{Reliability and Sensitivity}
\label{sec:Sensitivity-Analysis}
\label{sec:controlled}
\noindent\textbf{Ranking Uncertainty.}
We compute 10,000 paired bootstrap resamples of the 1,284 test images, sharing sampled indices across models. Table~\ref{tab:bootstrap-top} reports representative differences, and Table~\ref{tab:ranking-length} provides individual model intervals. The Claude Opus 4.6--Qwen3-VL-Plus and Qwen3-VL-32B--GPT-5.4 difference intervals include zero. Across all 68 models, 57 of the 67 unadjusted 95\% intervals for adjacent differences include zero. Thus, the leaderboard supports broad performance comparisons, while many small adjacent gaps remain unresolved. These intervals quantify image-sampling uncertainty under the evaluated pipeline.
\begin{table*}[t]
\centering
\small
\begin{tabular}{lrr}
\toprule
Comparison & Mean SIM difference & Paired bootstrap 95\% CI \\
\midrule
Claude Opus 4.6 -- Qwen3-VL-Plus & 0.0055 & [$-0.0013$, $0.0126$] \\
Qwen3-VL-Plus -- Qwen3-VL-Flash & 0.0066 & [$0.0002$, $0.0127$] \\
Qwen3-VL-Flash -- Qwen3-VL-32B & 0.0134 & [$0.0067$, $0.0201$] \\
Qwen3-VL-32B -- GPT-5.4 & 0.0006 & [$-0.0072$, $0.0083$] \\
GPT-5.4 -- GPT-5.2 & 0.0079 & [$0.0010$, $0.0148$] \\
\bottomrule
\end{tabular}
\caption{Representative paired differences on MMGenBench-Test using 10,000 image-level bootstrap resamples. Intervals are unadjusted for multiple comparisons. Differences use full-precision scores; displayed precision can cause small discrepancies with Table~\ref{tab:main-results-all}.}
\label{tab:bootstrap-top}
\end{table*}\space{}

\noindent\textbf{Controlled Factual Corruptions.}
To examine generator confounding, we use 150 image--attribute examples, with 30 each for Color, Geometry, Position, Count, and Text. Each example has a correct caption $C_+$ and a single-fact corruption $C_-$, reconstructed with matched settings using FLUX and SD3.5, giving 300 example--generator cases. We test whether $S(C_+)>S(C_-)$, where $S$ is reconstruction SIM for the fixed original image. A controllable subset includes only cases where both GPT-5.5 and Claude Opus 4.8 independently confirm, without access to SIM, that the intended fact in $C_+$ and the corrupted fact in $C_-$ are rendered.
\begin{table}[t]
\centering
\small
\begin{tabular}{lrr}
\toprule
Encoder & All cases & Controllable subset \\
 & ($n=300$) & ($n=219$) \\
\midrule
Unicom & 71.3 & 83.1 \\
CLIP & 72.0 & 81.7 \\
\bottomrule
\end{tabular}
\caption{Percentage of example--generator cases satisfying $S(C_+)>S(C_-)$ for correct and single-fact-corrupted captions. Two automatic verifiers select the controllable subset without accessing SIM.}
\label{tab:factual-sensitivity}
\end{table}\space{}
Of the 300 cases, 219 pass this verification; controllability rates are 74.7\% for FLUX and 71.3\% for SD3.5. Unicom distinguishes correct and corrupted captions in 71.3\% of all cases and 83.1\% of controllable cases (Table~\ref{tab:factual-sensitivity}). Across the two generators and two encoders, fact-preserving paraphrases yield a mean absolute score difference of $0.0073$, with 73.8\% within $0.01$. The verified subset shows stronger factual discrimination, consistent with generator controllability affecting the metric. These results support factual sensitivity on the automatically verified subset under the protocol in Appendix~\ref{appendix:controlled}.

Both generators and both encoders show higher discrimination on their controllable subsets. Omission variants yield positive correct-minus-omission mean differences, although a generator can supply omitted facts from its prior. Appendix~\ref{appendix:controlled} details selection, variants, and attribute-level results.

\noindent\textbf{Component and Seed Sensitivity.}
Across the tested generators and encoders, average SIM ranking correlations are high (Spearman $0.98$, Kendall $0.90$ for each sweep). Average per-model SIM standard deviation across generators is $0.013$. Five-seed evaluation of 23 models yields SIM standard deviations of $0.0002$--$0.0038$ (Appendix~\ref{appendix:Seed-Variance-of-Results}). These findings support stable rankings and low seed variation within the tested configurations; shared component failures remain possible. Full component results appear in Appendices~\ref{sec:Sensitivity-Analysis-of-T2I-Models} and~\ref{sec:Sensitivity-Analysis-of-Embedding-Models}.

\subsection{Caption Length Effects}
\label{sec:length-bias}
Across 68 models and 87,312 test captions, mean and median lengths are 51.19 and 39 words (IQR 25--56). Table~\ref{tab:main-results-all} reports mean lengths for the listed models; Table~\ref{tab:ranking-length} gives all 68 models. Model-level Spearman correlations of mean and median length with SIM are $0.248$ and $0.447$, respectively; the median per-image correlation is $0.239$. These associations capture length sensitivity without separating verbosity from added factual detail.

\noindent\textbf{Controlled Length Instructions.}
Table~\ref{tab:length-bias-summary} compares four models under $<20$, 20--60, and $>60$ word instructions. Every model's SIM increases as the requested description becomes longer. For example, InternVL2-8B increases from $0.4968$ to $0.5478$ and then $0.5701$.
\begin{table}[t]
\centering
\small
\setlength{\tabcolsep}{4pt}
\begin{tabular}{lccc}
\toprule
Model & SIM$_{<20}$ & SIM$_{20\text{-}60}$ & SIM$_{>60}$ \\
\midrule
InternVL2-8B         & 0.4968 & 0.5478 & 0.5701 \\
Ovis1.6-Gemma2-9B    & \textbf{0.5262} & \textbf{0.5825} & \textbf{0.5932} \\
Qwen2-VL-7B-Instruct & 0.4911 & 0.5323 & 0.5602 \\
LLaVA-OV-7B          & 0.4494 & 0.4888 & 0.5245 \\
\bottomrule
\end{tabular}
\caption{SIM-Score under three requested length constraints ($<\!20$, $20$--$60$, and $>\!60$ words). All four models retain the same relative ordering under the three requested description-length conditions.}
\label{tab:length-bias-summary}
\end{table}\space{}
All four models retain their relative ordering, while absolute SIM increases with requested length. Longer descriptions can change both factual detail and wording; the experiment captures their combined effect rather than verbosity alone. We retain the common 20--60 word instruction as the evaluation protocol and report actual lengths alongside the scores. Appendix~\ref{appendix:Length-Bias} gives the observed length distributions, including departures from the requested ranges.

\subsection{Computational Cost}
\label{sec:cost}
On one H20 with text-to-image batch size one, reconstruction takes 35.98 seconds per sample, feature extraction 0.0184 seconds, and metric computation an amortized 0.0064 seconds. Table~\ref{tab:cost} gives the resulting per-LMM budgets. Eight GPUs \mbox{reduce} wall-clock time through data parallelism; a single reconstruction uses one GPU. These figures exclude LMM caption generation and API charges. The protocol trades reference-annotation effort for generation compute, with a budget that scales with the image collection.
\begin{table}[t]
\centering
\small
\setlength{\tabcolsep}{4pt}
\begin{tabular}{lrrr}
\toprule
Set & Images & GPU-h & 8-GPU h \\
\midrule
Test & 1,284 & 12.84 & 1.61 \\
Domain & 10,000 & 100.01 & 12.50 \\
Combined & 11,284 & 112.86 & 14.11 \\
\bottomrule
\end{tabular}
\caption{Per-LMM evaluation cost on H20, excluding caption generation. GPU-hours are calculated from the displayed per-sample timings; eight-GPU wall-clock estimates assume ideal data parallelism.}
\label{tab:cost}
\end{table}\space{}

Stored per-image SIM supports model comparisons, pattern summaries, and paired resampling without repeating generation. Scores computed under a fixed configuration can be reused across caption pairings. New descriptions or generation settings require new reconstructions; encoder changes require recomputing features and scores.\space{}
\FloatBarrier
\section{Conclusion}
\label{sec:conclusion}
MMGenBench links reusable reconstruction scores to inspectable visual evidence, enabling comparison and diagnosis of LMM descriptions from raw images. Its two benchmark subsets combine distribution-level comparison with diagnosis across 13 overlapping visual patterns, linking overall performance to the visual content preserved within each pattern. Evaluation of 68 LMMs reveals differences among model variants and pattern profiles that aggregate scores can obscure. Inspecting descriptions and reconstructions links these differences to output behaviors, while stored scores support comparisons and paired resampling without repeating generation. The six-model human study reaches 90.00\% sample agreement; controlled factual changes support sensitivity on the verified subset, and component sweeps show stable rankings across the tested configurations. Project page: \href{https://github.com/lerogo/MMGenBench}{github.com/lerogo/MMGenBench}.\space{}
\clearpage
\section*{Limitations}
MMGenBench measures the preservation of reconstructible visual information under a fixed generator and encoder. Its scores reflect description content, instruction following, generator controllability, and encoder sensitivity. Correct descriptions can receive low scores when text, counts, or spatial relations cannot be rendered or distinguished reliably. Component agreement does not exclude shared failures. The controllable-subset analysis is conditional on automatic verification and may favor easier examples.

Both main benchmarks derive from JourneyDB, while the four-model COCO experiment provides initial evidence for natural photographs. Applying the framework to documents, screenshots, or medical images requires separate validation and potentially specialized rendering or comparison models. Human preferences cover six LMMs, rather than all 68 models. Bootstrap intervals quantify image-sampling uncertainty under the evaluated configuration and exclude API-version and generation variability. The SIM-selected empirical envelope is an anchor within the current model pool, not a human or theoretical ceiling.

Caption length affects absolute scores, and the four-model length experiment does not establish invariance to verbosity. Small-sample Fr\'echet distances remain exploratory. Reconstruction requires substantial computation despite avoiding reference annotation; the reported budget excludes caption generation and API charges. Results are conditional on the evaluated model versions and pipeline configuration. Comparisons across changed generators, encoders, or API versions require consistent evaluation settings.

\section*{Ethical Considerations}
MMGenBench is a diagnostic resource for visual description, rather than a measure of model safety or societal impact. Its image collections, generators, and encoders may inherit representational biases that affect scores across visual domains. Benchmark use should respect the licenses and intended uses of the source artifacts. The JourneyDB human preference study used nine unpaid volunteer computer science master's students from the authors' and collaborating laboratories; their training and annotation task are described in Appendix~\ref{appendix:Human-Alignment-Score}. Their judgments should not be treated as representative of all user populations.\space{}

\clearpage
\appendix
\clubpenalty=10000
\widowpenalty=10000
\displaywidowpenalty=10000

\begin{figure*}[!ht]
\centering
\includegraphics[width=1.\textwidth]{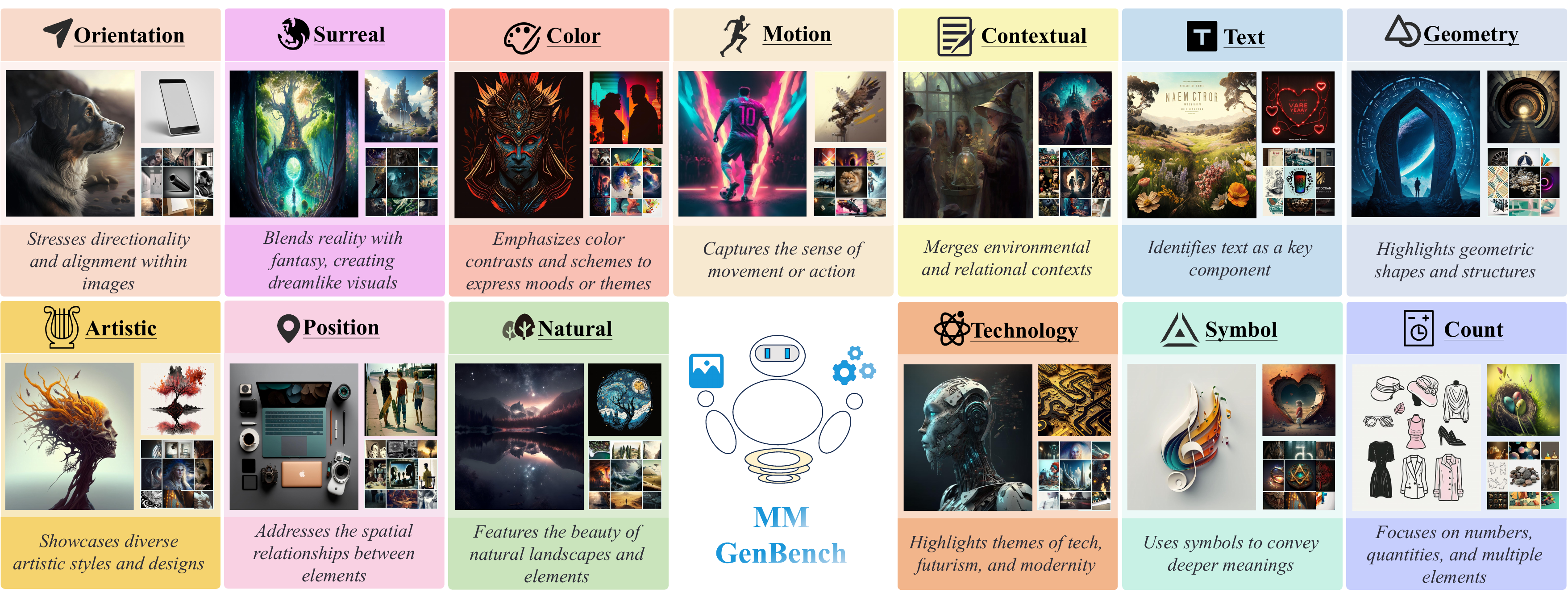}
\caption{Examples from MMGenBench-Test. The dataset contains $13$ distinct image patterns, and each pattern is illustrated with representative images and a concise textual explanation.}
\label{fig:dataset-case}
\end{figure*}

\space{}

\section{Details of MMGenBench Benchmark}
\label{appendix:MMGenBench-Benchmark-Details}
\begin{figure*}[t]
\centering
\includegraphics[width=1.\textwidth]{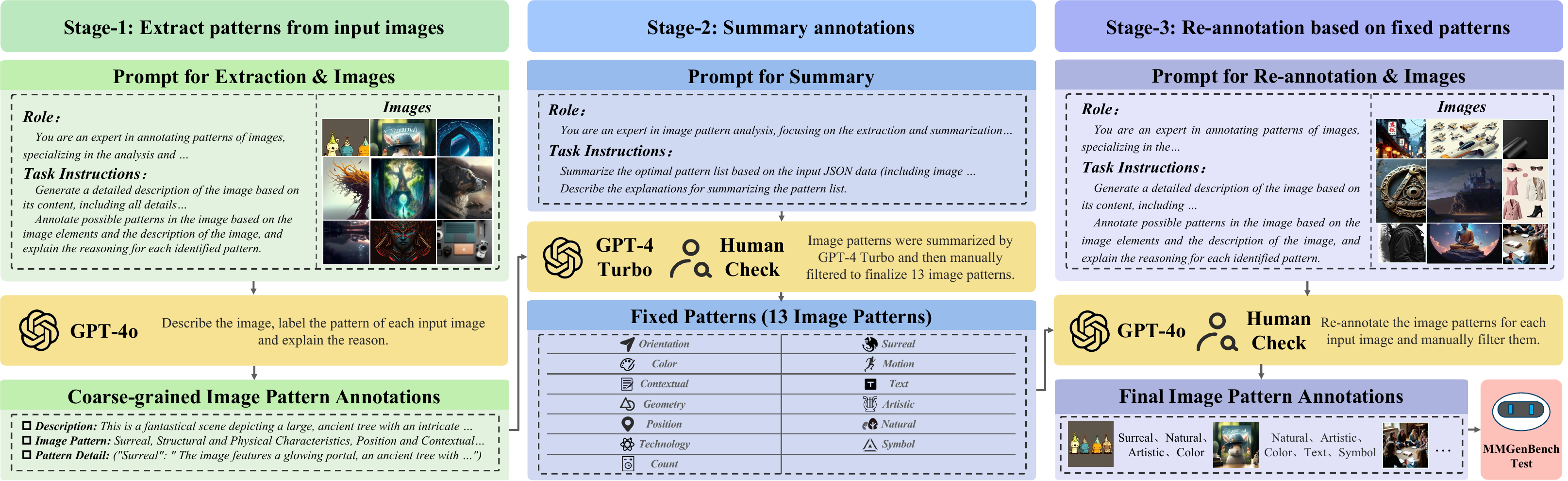}

\caption{Overview of the MMGenBench-Test construction process. We use GPT-4o to extract image patterns from input images, GPT-4 Turbo to summarize the extracted patterns, and human verification to finalize $13$ image patterns. GPT-4o is then used for re-annotation, followed by human review to produce the final annotations.}
\label{fig:MMGenBench-test}
\end{figure*}\space{}
\subsection{Overview}
MMGenBench-Pipeline scores LMM descriptions directly from raw images, enabling automated evaluation without per-image questions or reference captions. Building on this property, we first construct MMGenBench-Domain, a domain-level dataset in the generated-image domain, to evaluate the visual information preserved in LMM descriptions of generated images; it is built by sampling images from the JourneyDB~\cite{sun2024journeydb} validation set with no additional annotation. To examine these descriptions across diverse image types, we additionally construct MMGenBench-Test, a curated test set covering $13$ image patterns from the JourneyDB test set. We use a multi-stage procedure to extract and annotate image patterns, as illustrated in Fig.~\ref{fig:MMGenBench-test}, and manually double-check the image patterns before finalizing the annotations. Beyond these two datasets, MMGenBench-Pipeline can be applied to other domains by replacing the input image collection with images from the target domain.

\subsection{Dataset Statistics}
MMGenBench-Domain is constructed by randomly sampling $10{,}000$ images from the JourneyDB validation set, which allows us to evaluate the reconstructible information in LMM descriptions within the generated-image domain without any additional annotation. MMGenBench-Test contains $1{,}284$ images across $13$ distinct image patterns (see Appendix~\ref{appendix:MMGenBench-Test-Data-Details}). The distribution of images per pattern is shown in Fig.~\ref{fig:sample_statistics}, where each pattern contains at least $114$ images. An image may contain multiple patterns; for instance, the first image annotation in Fig.~\ref{fig:MMGenBench-test} contains four image patterns: ``Surreal'', ``Natural'', ``Artistic'' and ``Color''. These $13$ image patterns support diagnosis of reconstructible information across diverse visual dimensions, and all samples are manually checked and annotated to ensure overall quality.

\subsection{MMGenBench-Domain Construction}
\label{appendix:MMGenBench-Domain-Details}
Constructing high-quality datasets typically requires substantial annotation effort. Therefore, efficient methods for creating domain-specific evaluation sets are important when assessing LMM descriptions in new domains. MMGenBench-Pipeline can score images from different domains, although metric validity must be established separately in each domain. In this study, we randomly select $10{,}000$ images from the JourneyDB validation set to construct MMGenBench-Domain. We use SIM-Score as the primary metric and FID-Score as a secondary distributional metric.

\subsection{MMGenBench-Test Data Collection}
\label{appendix:MMGenBench-Test-Data-Details-Collection}
\noindent\textbf{Pattern Extraction.}
We extract image patterns from existing images to analyze LMM descriptions across various image categories. Because the construction setting assumes only image inputs are available, we use GPT-4o to extract and analyze underlying image patterns. The task is executed by GPT-4o in three sequential steps: 1) providing a detailed description of the image; 2) annotating the possible patterns based on the image features and description; and 3) explaining the rationale behind the annotated image patterns.
With the prompts described in Appendix \ref{appendix:Construction-Prompts-Details}, GPT-4o produces image descriptions, pattern annotations, and rationales for the target images. An example is presented in Fig. \ref{fig:pattern-extraction-case} in Appendix \ref{appendix:Image-Pattern-Extraction-Case}, where the image description, extracted image patterns, and rationales are consistent with the image content. By annotating each image in the JourneyDB test set, we obtain a total of $1,868$ distinct pattern labels.

\noindent\textbf{Annotation Summary.}
By leveraging GPT-4o, we extract a broad set of image patterns and observe that some patterns overlap semantically. We compute frequency statistics for the extracted patterns and rank them in descending order (as shown in Appendix \ref{appendix:MMGenBench-Test-Data-Details}). We then use GPT-4-Turbo to summarize these patterns with the prompt shown in Appendix \ref{appendix:Construction-Prompts-Details}. This procedure results in a curated list of $13$ distinct image patterns, each described in Appendix \ref{appendix:MMGenBench-Test-Data-Details}.

\noindent\textbf{Image Re-annotation.}
We re-annotate the JourneyDB test set using the previously identified $13$ image patterns. Similar to the pattern extraction stage, GPT-4o is used to generate descriptions for each image. These descriptions are then annotated with the predefined image patterns, together with the corresponding rationales. A key difference in this stage is that re-annotation is restricted to the specified $13$ image patterns and excludes other patterns. Given the size of the test set, we select a subset of images for subsequent processing. The re-annotated images are classified into distinct image patterns, each containing a specified number of images. To construct the final dataset, we randomly sample a subset of images from each pattern with a selection probability of $\frac{100}{N}$, where $N$ represents the total number of images within the pattern. To mitigate potential annotation errors, we manually verify and annotate each image. Model suggestions and manual annotations are consolidated into the final multi-label assignments. The resulting patterns serve as manually checked organizational tags for diagnosis. An image can belong to multiple sampled patterns. The final set contains 1,284 unique images with their consolidated multi-label assignments; overall scores weight each image once, independently of its number of labels. The 100/$N$ sampling probability targets coverage within each candidate pattern, rather than a final quota of exactly 100 images after overlapping memberships are consolidated.

\subsection{Details of MMGenBench-Test Data}
\label{appendix:MMGenBench-Test-Data-Details}
\begin{figure*}[ht]
\centering
\begin{tcolorbox}[title=13 Image Patterns]
\large
\includegraphics[width=0.3cm]{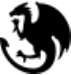} Surreal
\small
\\
This pattern is characterized by its prevalence in depicting scenes that mix elements of fantasy with reality, often creating imaginative or dream-like visuals.
\\
\large
\includegraphics[width=0.35cm]{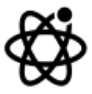} Technology
\small
\\
Highlights themes related to technology, futurism, and modernity, encompassing various aspects of technological advances and speculative futures.
\\
\large
\includegraphics[width=0.35cm]{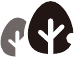} Natural
\small
\\
Encompasses imagery related to natural sceneries or elements, drawing on the beauty and complexity of the \mbox{natural world}.
\\
\large
\includegraphics[width=0.3cm]{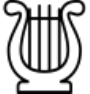} Artistic
\small
\\
Captures various artistic styles and decorations, reflecting creativity, design elements, and unique art styles.
\\
\large
\includegraphics[width=0.35cm]{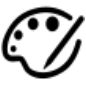} Color
\small
\\
Focuses on the use and significance of color in images, including aspects like color contrast, schemes, and gradients to convey moods or themes.
\\
\large
\includegraphics[width=0.3cm]{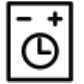} Count
\small
\\
Deals with patterns that emphasize numerical aspects, quantities, and the presence of multiple elements, indicating a focus on enumeration or amount.
\\
\large
\includegraphics[width=0.3cm]{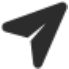} Orientation
\small
\\
Emphasizes the directionality and alignment within images, capturing how subjects or elements are oriented or directed.
\\
\large
\includegraphics[width=0.25cm]{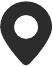} Position
\small
\\
Concerns the relative placement and spatial relationships between elements within an image, highlighting how objects are positioned.
\\
\large
\includegraphics[width=0.3cm]{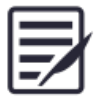} Contextual
\small
\\
Fuses themes of environmental and relational context, shedding light on the settings and the interconnections \mbox{within images}.
\\
\large
\includegraphics[width=0.3cm]{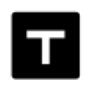} Text
\small
\\
Identifies patterns where text is a significant component, showcasing how textual content contributes to the overall message or theme of the image.
\\
\large
\includegraphics[width=0.35cm]{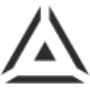} Symbol
\small
\\
Centers on the use of symbols and symbolic elements to convey deeper meanings or concepts, drawing from various symbolic traditions.
\\
\large
\includegraphics[width=0.35cm]{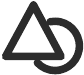} Geometry
\small
\\
Illustrates a focus on geometric shapes, patterns, and arrangements, underlining the role of structure and form in images.
\\
\large
\includegraphics[width=0.35cm]{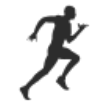} Motion
\small
\\
Captures the sense of movement or dynamics within images, depicting actions or the concept of motion through \mbox{visual elements}.
\end{tcolorbox}
\caption{The $13$ image patterns in MMGenBench-Test, obtained through GPT-4 Turbo summarization and \mbox{human verification}.}
\label{tc:13-image-patterns}
\end{figure*}

\noindent\textbf{Thirteen Image Patterns.}
As illustrated in Fig. \ref{fig:dataset-case} and \ref{tc:13-image-patterns}, summarization is performed using GPT-4 Turbo, followed by human verification. We identify $13$ distinct image patterns, each with a detailed explanation. A single image may exhibit multiple patterns concurrently.

\noindent\textbf{Patterns Extracted by GPT-4o.}
Fig. \ref{tc:extracted-image-patterns} shows the image patterns extracted by GPT-4o, organized in descending order. While the number of distinct extracted pattern labels is 1,868, only the most frequently occurring patterns are presented due to space constraints.

\begin{figure}[t]
\centering
\includegraphics[width=0.8\linewidth]{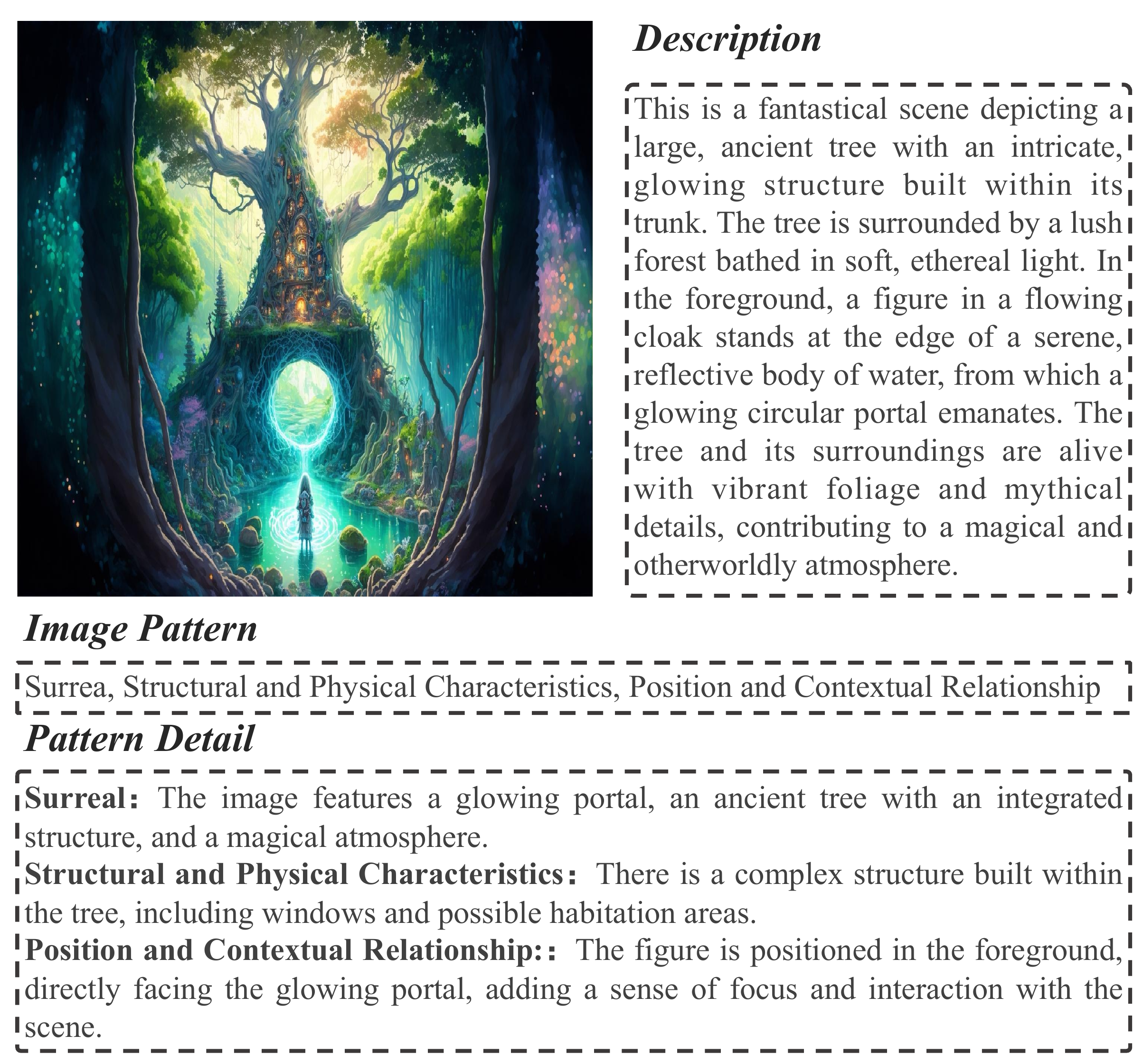}
\caption{Case study of image pattern extraction. GPT-4o generates an image description, identifies underlying image patterns, and explains the extracted patterns.}
\label{fig:pattern-extraction-case}
\end{figure}\space{}
\subsection{Case Study of Pattern Extraction}
\label{appendix:Image-Pattern-Extraction-Case}
Fig. \ref{fig:pattern-extraction-case} illustrates an example in which GPT-4o identifies image patterns from a given image. The results show clear correspondence among the image description, the extracted patterns, and the corresponding explanations, all of which are consistent with the visual content.

\subsection{Details of Construction Prompts}
\label{appendix:Construction-Prompts-Details}
To construct MMGenBench-Test, we design a series of prompts tailored to the requirements of each stage. Fig. \ref{tc:Prompt-for-Extraction}, \ref{tc:Prompt-for-Summary}, and \ref{tc:Prompt-for-Re-annotation} correspond to the extraction, summarization, and re-annotation stages in Fig.~\ref{fig:MMGenBench-test}.

\section{Additional Experimental Details}
\label{appendix:Additional-Experimental-Details}
\subsection{Evaluation Pipeline Prompt}
\label{appendix:Evaluation-Pipeline-Prompt}
We design a prompt to guide LMMs in generating textual descriptions of images for subsequent evaluation. As shown in Fig. \ref{tc:Evaluation-Pipeline-Prompt}, the prompt covers five perspectives: ``role'', ``definition'', ``task'', ``key points'', and ``output''. To maintain evaluation consistency across models, we use the same prompt for all evaluated LMMs.

\subsection{Experimental Setup}
\label{appendix:Experimental-Setup}
We evaluate 68 representative LMMs, as detailed in Tables~\ref{tab:suppl-results-all} and~\ref{tab:suppl-results-test}, with evaluation infrastructure from OpenCompass~\cite{2023opencompass}. For open-source LMMs, we use the original configuration of VLMEvalKit~\cite{duan2024vlmevalkit} to generate textual descriptions of images on $8$ NVIDIA-H20 GPUs. During the image generation and metric calculation stages, we also use $8$ NVIDIA-H20 GPUs for multi-process inference. The default reconstruction model is FLUX.1-dev and the encoder is Unicom/ViT-L-14-336px. The evaluation records contain the 87,312 Test captions, source-image IDs, per-image SIM values, and bootstrap summaries. These records support model-level, pattern-level, and caption-length analyses without regenerating images. Human-study records additionally retain the nine votes for each comparison. Reusing a score across caption pairings leaves its original image, description, and reconstruction unchanged. Table~\ref{tab:protocol-summary} brings together the evaluation and validation protocols. For implementation details, see \href{https://github.com/lerogo/MMGenBench}{github.com/lerogo/MMGenBench}.
\begin{table*}[t]
\centering
\small
\setlength{\tabcolsep}{5pt}
\begin{tabular}{p{0.23\textwidth}p{0.70\textwidth}}
\toprule
Stage or study & Protocol \\
\midrule
Description generation & VLMEvalKit model configurations with the common prompt in Fig.~\ref{tc:Evaluation-Pipeline-Prompt}; requested length 20--60 words. Actual lengths are reported separately. \\
Default reconstruction & FLUX.1-dev; eight H20 GPUs provide data parallelism, with one GPU per reconstruction. \\
Representation and score & Unicom/ViT-L-14-336px; cosine similarity between original and reconstructed image embeddings. Benchmark scores average per-image SIM. \\
JourneyDB human study & Six LMMs, all 15 unordered model pairs, 1,000 comparisons, and nine independent votes per comparison. A sample decision requires at least five supporting votes. \\
SIM preference rule & Caption~1 if $\Delta>0.01$, Caption~2 if $\Delta<-0.01$, and comparable otherwise; the threshold is fixed before examining human labels. \\
VLM judges & GPT-5.5 and Claude Opus 4.8, zero-shot judging in both caption orders; reversed predictions are mapped back before averaging agreement. \\
Factual perturbations & Correct, corrupted, omitted, and paraphrased variants across five attributes; FLUX and SD3.5 use matched generation settings. \\
Controllability checks & Both GPT-5.5 and Claude Opus 4.8 must confirm that the respective target facts in both reconstructions are rendered, without access to SIM. \\
Ranking uncertainty & 10,000 paired image-level bootstrap resamples; each replicate uses the same sampled Test indices for every model. \\
\bottomrule
\end{tabular}
\caption{Summary of the evaluation and validation protocols. Human-study and factual-perturbation details appear in Appendices~\ref{appendix:Human-Alignment-Score} and~\ref{appendix:controlled}.}
\label{tab:protocol-summary}
\end{table*}\space{}

\subsection{Comprehensive Experimental Evaluation}
\label{appendix:Detailed-Experiment}
\begin{figure}[htbp]
\centering
\includegraphics[width=\linewidth]{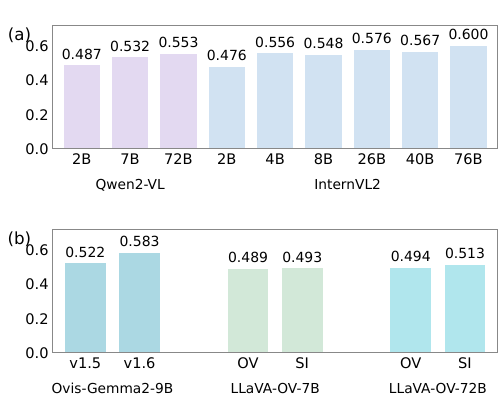}
\caption{SIM comparison of representative LMMs on MMGenBench-Test: (a) parameter sizes within Qwen2-VL and InternVL2; (b) Ovis and LLaVA-OneVision variants. These descriptive comparisons highlight score differences across the evaluated model sizes and variants.}
\label{fig:case-model-cmp}
\end{figure}

Results for all 68 LMMs appear in Tables~\ref{tab:suppl-results-all} and~\ref{tab:suppl-results-test}. Figure~\ref{fig:case-model-cmp} compares selected model \mbox{sizes and variants}.
\space{}
\begin{table*}[t]
\centering
\small
\setlength{\tabcolsep}{4pt}
\begin{tabular}{lrrrrrrrr}
\toprule
Pattern & Claude & GPT & Qwen & Qwen & InternVL & MiniCPM & LLaVA & Ovis \\
 & Opus 4.6 & 5.4 & 3-VL-Plus & 3-VL-32B & 3.5-38B & V2.6 & OV-SI-72B & 1.6-9B \\
\midrule
Color & 0.667 & 0.645 & 0.663 & 0.648 & 0.600 & 0.559 & 0.530 & 0.601 \\
Position & 0.651 & 0.622 & 0.639 & 0.620 & 0.556 & 0.517 & 0.477 & 0.547 \\
Contextual & 0.636 & 0.611 & 0.617 & 0.600 & 0.530 & 0.492 & 0.457 & 0.518 \\
Text & 0.617 & 0.588 & 0.652 & 0.631 & 0.585 & 0.524 & 0.500 & 0.573 \\
Surreal & 0.668 & 0.649 & 0.657 & 0.644 & 0.594 & 0.565 & 0.540 & 0.593 \\
Artistic & 0.671 & 0.646 & 0.664 & 0.645 & 0.594 & 0.559 & 0.518 & 0.593 \\
Orientation & 0.649 & 0.635 & 0.627 & 0.601 & 0.541 & 0.500 & 0.467 & 0.540 \\
Geometry & 0.658 & 0.618 & 0.662 & 0.630 & 0.586 & 0.577 & 0.512 & 0.595 \\
Technology & 0.653 & 0.635 & 0.654 & 0.627 & 0.576 & 0.540 & 0.517 & 0.579 \\
Count & 0.663 & 0.633 & 0.653 & 0.644 & 0.579 & 0.547 & 0.515 & 0.568 \\
Natural & 0.675 & 0.662 & 0.667 & 0.648 & 0.592 & 0.558 & 0.527 & 0.580 \\
Symbol & 0.698 & 0.665 & 0.695 & 0.668 & 0.625 & 0.594 & 0.559 & 0.633 \\
Motion & 0.659 & 0.644 & 0.646 & 0.636 & 0.573 & 0.522 & 0.509 & 0.576 \\
\bottomrule
\end{tabular}
\caption{Per-pattern SIM for the eight models in Fig.~\ref{fig:image-pattern-result-cmp}. Ovis denotes Ovis1.6-Gemma2-9B. The radar chart uses $100\times$SIM; this table uses unscaled SIM. These means provide descriptive performance profiles over the overlapping image subsets.}
\label{tab:pattern-selected}
\end{table*}\space{}

\subsection{Sensitivity Analysis of T2I Models}
\label{sec:Sensitivity-Analysis-of-T2I-Models}
We conducted a sensitivity analysis on four text-to-image models (FLUX.1-dev, SD3.5, Kolors, Lumina). Correlations compare model-level LMM rankings obtained from mean Test scores under each generator. The results show:

\noindent(1) \textbf{LMM rankings show high cross-generator agreement.} SIM correlations across the four generators are high (Spearman~$=0.98$, Kendall~$=0.90$ on average). This indicates that changing the T2I model has limited effect on the relative ranking of LMMs (Tables~\ref{sensitivity_analysis_of_t2i_model:1},~\ref{sensitivity_analysis_of_t2i_model:2}).

\noindent(2) \textbf{FID rankings show lower cross-generator agreement than SIM.} FID correlations remain high (Spearman~$=0.91$, Kendall~$=0.78$), indicating greater ranking sensitivity to generator choice than SIM (Tables~\ref{sensitivity_analysis_of_t2i_model:3},~\ref{sensitivity_analysis_of_t2i_model:4}).

\noindent(3) \textbf{Absolute score sensitivity varies across LMMs.} Table~\ref{sensitivity_analysis_of_t2i_model:6} lists the largest cross-generator SIM standard deviations and score ranges. These measure absolute score variation, complementing the ranking correlations above.

\noindent(4) \textbf{Absolute SIM variation is modest across the tested generators.} The average cross-generator standard deviation is $0.013$, and the average range is $0.03$ (Table~\ref{sensitivity_analysis_of_t2i_model:7}).

Together, these results support ranking stability and modest SIM variation across the four tested generators. Factual validity requires separate evidence because generators may share failure modes.

\subsection{Sensitivity Analysis of Embedding Models}
\label{sec:Sensitivity-Analysis-of-Embedding-Models}
We conduct a sensitivity analysis across three widely used embedding models (Unicom/ViT-L-14-336px~\cite{unicom}, DINOv2/dinov2-large~\cite{dinov2}, CLIP/clip-vit-large-patch14-336~\cite{clip}). The results (Tables~\ref{sensitivity_analysis_of_embedding_model:1}--\ref{sensitivity_analysis_of_embedding_model:4}) show high agreement of LMM rankings across the three tested encoders. SIM-Scores exhibit high cross-model correlations (Spearman~$=0.98$, Kendall~$=0.90$), and FID-Scores are even more consistent (Spearman~$=0.99$, Kendall~$=0.92$). These results support ranking stability across Unicom, DINOv2, and CLIP; shared insensitivity to particular facts remains possible.

\subsection{Length Bias of LMM}
\label{appendix:Length-Bias}
We conducted a controlled study on four representative LMMs
(InternVL2-8B, Ovis1.6-Gemma2-9B, Qwen2-VL-7B-Instruct, LLaVA-OneVision-Qwen2-7B) under three explicit word-count constraints:
$<20$, $20-60$ (our default), and $>60$ words~(Table~\ref{addtab:word_count_stats}).

As shown in Table~\ref{addtab:length_metrics}, across all models, SIM increases and FID decreases monotonically with requested length
(e.g., InternVL2-8B: SIM $0.4968 \rightarrow 0.5478 \rightarrow 0.5701$; Ovis1.6-Gemma2-9B: $0.5262 \rightarrow 0.5825 \rightarrow 0.5932$), indicating that very short captions ($<20$ words) may omit essential semantics. The ordering of these four models remains unchanged across the three instructions, but the actual length distributions differ substantially between models. The experiment varies requested length rather than holding factual content fixed; it therefore cannot isolate a pure verbosity effect. The 20--60 word instruction provides a common target; Table~\ref{addtab:word_count_stats} reports actual lengths and departures from that target. Across all 68 models and 87,312 captions, mean/median lengths are 51.19/39 words (IQR 25--56). Model-level Spearman correlations between mean/median length and SIM are 0.248/0.447, and the median per-image correlation is 0.239. Table~\ref{tab:ranking-length} provides lengths and SIM intervals for every model. These 68-model summaries count words by whitespace separation. Their descriptive correlations do not isolate the effects of verbosity and \mbox{factual detail}.

\begin{table*}[!th]
\centering
\small
\begin{tabular}{lcccccc}
\toprule
Model & $<20$ Mean & $<20$ Std & 20-60 Mean & 20-60 Std & $>60$ Mean & $>60$ Std \\
\midrule
InternVL2-8B             & 19.17 &  4.70 & 38.87 & 12.51 & 81.81 & 83.38 \\
Ovis1.6-Gemma2-9B        & 22.49 &  6.55 & 55.85 & 12.26 & 90.43 & 18.39 \\
Qwen2-VL-7B-Instruct     & 15.48 & 21.06 & 27.67 &  9.08 & 68.36 & 66.71 \\
LLaVA-OneVision-Qwen2-7B & 11.09 &  3.37 & 17.35 &  6.99 & 36.75 & 24.88 \\
\bottomrule
\end{tabular}
\caption{Word-count statistics from the four-model controlled length study.}
\label{addtab:word_count_stats}
\end{table*}

\begin{table*}[!th]
\centering
\small
\begin{tabular}{lcccccc}
\toprule
Model & $<20$ SIM & $<20$ FID & 20--60 SIM & 20--60 FID & $>60$ SIM & $>60$ FID \\
\midrule
InternVL2-8B             & 0.4968 & 1.5337 & 0.5478 & 1.4037 & 0.5701 & 1.3171 \\
Ovis1.6-Gemma2-9B        & 0.5262 & 1.4545 & 0.5825 & 1.3167 & 0.5932 & 1.2645 \\
Qwen2-VL-7B-Instruct     & 0.4911 & 1.5562 & 0.5323 & 1.4370 & 0.5602 & 1.3651 \\
LLaVA-OneVision-Qwen2-7B & 0.4494 & 1.7247 & 0.4888 & 1.5904 & 0.5245 & 1.4734 \\
\bottomrule
\end{tabular}
\caption{SIM-Score and FID-Score under different requested length constraints. FID estimates on MMGenBench-Test and its patterns are exploratory.}
\label{addtab:length_metrics}
\end{table*}\space{}

\subsection{Human Alignment Score}
\label{appendix:Human-Alignment-Score}
\noindent\textbf{Comparison Sampling.}
We select six representative LMMs: GPT-4o-high, InternVL2-76B, Qwen2-VL-72B-Instruct, LLaVA-OneVision-Qwen2-72B, MiniCPM-V-2.6, and Ovis1.6-Gemma2-9B. For each comparison, we randomly select two models and one MMGenBench-Test image and present the two corresponding descriptions. The study contains 1,000 pairs from 702 distinct source images and covers all 15 unordered model pairs (Table~\ref{tab:human-pair-counts}). The main human-alignment analysis and baseline comparison reuse these same pairs.

\noindent\textbf{Annotators and Instructions.}
Nine volunteer computer science master's students from our and collaborating laboratories independently annotate every comparison. All receive unified oral training, and no annotator is paid. Each sees the original image and the two descriptions and chooses C1 if Caption~1 describes the image better, C2 if Caption~2 describes it better, or B if they are comparable. A majority label requires at least five of nine votes. There are 493 C1 majorities, 450 C2 majorities, and 56 B majorities. One comparison has counts C1=2, C2=3, B=4 and hence no majority. All 1,000 comparisons and 9,000 votes are retained.

\noindent\textbf{MMGenBench Decision and Agreement.}
Let $\Delta=S(C_1)-S(C_2)$. The metric chooses C1 for $\Delta>0.01$, C2 for $\Delta<-0.01$, and B for $|\Delta|\leq0.01$. This threshold was fixed before inspecting human labels. Let $n_i(d)$ count annotators choosing decision $d$ for comparison $i$. Sample agreement is $\frac{1}{1000}\sum_i\mathbf{1}[n_i(d_i)\geq5]$ for metric decision $d_i$; the no-majority comparison contributes zero. Vote agreement is $\frac{1}{9000}\sum_i n_i(d_i)$. MMGenBench obtains 900/1,000 and 7,944/9,000, yielding 90.00\% and 88.27\%. Fleiss' $\kappa=0.940$ and nominal Krippendorff's $\alpha=0.940$ indicate high inter-annotator agreement. The nominal comparison-level Wilson 95\% interval is 88.0\%--91.7\%, a conditional summary of this study without adjustment for repeated source images or annotator recruitment. Individual votes are not independent trials.\begin{table*}[t]
\centering
\small
\begin{tabular}{llr}
\toprule
Model 1 & Model 2 & Comparisons \\
\midrule
GPT-4o-high & InternVL2-76B & 75 \\
GPT-4o-high & LLaVA-OV-72B & 62 \\
GPT-4o-high & MiniCPM-V-2.6 & 62 \\
GPT-4o-high & Ovis1.6-Gemma2-9B & 66 \\
GPT-4o-high & Qwen2-VL-72B & 48 \\
InternVL2-76B & LLaVA-OV-72B & 61 \\
InternVL2-76B & MiniCPM-V-2.6 & 73 \\
InternVL2-76B & Ovis1.6-Gemma2-9B & 69 \\
InternVL2-76B & Qwen2-VL-72B & 69 \\
LLaVA-OV-72B & MiniCPM-V-2.6 & 68 \\
LLaVA-OV-72B & Ovis1.6-Gemma2-9B & 81 \\
LLaVA-OV-72B & Qwen2-VL-72B & 59 \\
MiniCPM-V-2.6 & Ovis1.6-Gemma2-9B & 79 \\
MiniCPM-V-2.6 & Qwen2-VL-72B & 68 \\
Ovis1.6-Gemma2-9B & Qwen2-VL-72B & 60 \\
\midrule
Total & & 1,000 \\
\bottomrule
\end{tabular}
\caption{All 15 unordered model pairs in the archived JourneyDB human study. The 1,000 comparisons use 702 distinct source images and nine votes per comparison. A method agrees at the sample level only when at least five votes support its decision; the one comparison without any five-vote majority contributes zero to every method.}
\label{tab:human-pair-counts}
\end{table*}\space{}

\noindent\textbf{Score Margins and Scope.}
Human agreement increases with the absolute SIM margin, supporting the margin's use in interpreting pairwise decisions (Table~\ref{tab:human-margin}). Small margins remain difficult even when global model means differ. Human-alignment evidence covers these six models and their 15 pairings. Controlled perturbations assess factual sensitivity, while paired bootstrap analysis quantifies image-sampling uncertainty in the 68-model comparisons.

\noindent\textbf{Baselines.}
Direct CLIP compares each caption with its source image. The VFC-style reconstruction baseline applies CLIP to the same original and reconstructed images used by Unicom, holding the reconstructed inputs fixed. GPT-5.5 and Claude Opus 4.8 use a zero-shot pairwise judging protocol in both caption orders. After mapping reversed predictions back to the original caption identities, we compute sample and vote agreement for each order using the human votes and denominators above. Table~\ref{tab:comparison-with-clip} reports the arithmetic mean of the two agreement rates. A pair whose majority label is matched in only one order contributes one half to the averaged sample count, giving 0.05-percentage-point resolution over 1,000 pairs. Order consistency is the fraction of pairs with identical mapped decisions across the two orders, irrespective of human agreement; the judges change decisions on 67 and 95 pairs, respectively.

These comparisons characterize human alignment under the tested scoring and decision protocols. The zero-shot judges cover the evaluated configurations rather than the full CapArena-Auto evaluator, and the comparisons do not isolate encoder-only effects.

\begin{samepage}
For the direct CLIP comparison, 63 sample decisions are correct only under CLIP and 356 only under MMGenBench, consistent with their 607 versus 900 correct majority decisions. At the vote level, the corresponding counts are 660 and 3,109. These paired counts describe the observed agreement differences; repeated annotator votes are not independent observations for significance testing. The smaller Unicom--CLIP-reconstruction gap is also reported descriptively.
\par\end{samepage}

\subsection{Controlled Corruptions}
\label{appendix:controlled}
The study contains 150 image--attribute examples drawn from 137 distinct images, with 30 examples each for Color, Geometry, Position, Count, and Text. Images are selected using predefined salience and unambiguity criteria before metric computation. For each example, $C_+$ states the correct target fact, $C_-$ changes only that fact, $C_0$ omits it, and $C_{\mathrm{para}}$ preserves the facts while changing the wording. Both FLUX and SD3.5 reconstruct all four variants under matched settings. This design compares an explicit error, missing information, and a wording change against the same correct description. Section~\ref{sec:controlled} reports \mbox{correct-versus-corrupted discrimination}.

The controllability check concerns whether each reconstruction renders its specified fact, including the corrupted fact in $C_-$. GPT-5.5 and Claude Opus 4.8 independently inspect the reconstructions without observing SIM; both verifiers must confirm that both reconstructions render their specified facts. This retains 219 of 300 example--generator cases. Images recur across generators and occasionally across attributes, so rates describe cases rather than independent images. All-case rates include rendering failures, whereas subset rates condition on automatic verification, which may favor easier examples and inherit verifier biases. Their difference does not quantify the generator's \mbox{causal contribution}.

Table~\ref{tab:factual-by-attribute} presents attribute-level and aggregate results for each generator. Unicom prefers the correct caption in more than half of the cases in every attribute group under both generators. Its aggregate discrimination rises from 70.0\% to 81.3\% for FLUX and from 72.7\% to 85.0\% for SD3.5 on their controllable subsets. CLIP shows the same direction, with all-case rates of 73.3\% and 70.7\%. Both generators and both encoders support the pooled increase in discrimination. Sensitivity varies across attributes, and the aggregate trend allows differing decisions on individual cases.

Paraphrases pool 600 absolute score differences across 150 examples, two generators, and two encoders: the mean is 0.0073, and 73.8\% lie within 0.01. For omissions, mean $S(C_+)-S(C_0)$ is positive in all 20 attribute--generator--encoder settings. Across these tested settings, omitted facts lower SIM on average; a generator's prior may still supply missing facts.
\begin{table*}[t]
\centering
\small
\setlength{\tabcolsep}{5pt}
\begin{tabular}{lrrrr}
\toprule
Attribute & Controllability & Unicom: all & Unicom: controllable & CLIP: controllable \\
\midrule
Color & 76.7 / 70.0 & 63.3 / 76.7 & 73.9 / 90.5 & 78.3 / 76.2 \\
Geometry & 76.7 / 76.7 & 73.3 / 80.0 & 82.6 / 82.6 & 82.6 / 82.6 \\
Position & 76.7 / 63.3 & 73.3 / 60.0 & 82.6 / 89.5 & 82.6 / 84.2 \\
Count & 63.3 / 70.0 & 60.0 / 63.3 & 84.2 / 81.0 & 84.2 / 81.0 \\
Text & 80.0 / 76.7 & 80.0 / 83.3 & 83.3 / 82.6 & 83.3 / 82.6 \\
\midrule
All attributes & 74.7 / 71.3 & 70.0 / 72.7 & 81.3 / 85.0 & 82.1 / 81.3 \\
\bottomrule
\end{tabular}
\caption{Factual discrimination by attribute (\%). Entries give FLUX / SD3.5, with 30 image--attribute examples per attribute and generator; the last row pools 150 examples per generator. Discrimination is the fraction satisfying $S(C_+)>S(C_-)$. Controllability is the fraction retained by the two-verifier check. Rates describe cases, not independent images; source images may recur across attributes.}
\label{tab:factual-by-attribute}
\end{table*}\space{}

\subsection{Empirical Envelope and Ranking Uncertainty}
\label{appendix:empirical-envelope}
JourneyDB-Test does not provide original generation prompts. Validation prompts have mean/median lengths of 24.88/18 words (234,156 images; IQR 10--31), and 25.09/18 for the 10,000-image Domain subset. They may omit details realized during synthesis and are not \mbox{detailed-caption oracles}.

The empirical envelope in Eq.~\ref{equ:envelope} selects the highest SIM among 68 descriptions per test image, reaching $0.768$ versus $0.663$ for the best single model. Reusing SIM for selection and evaluation makes this a candidate-pool summary, not a human-caption or generator ceiling.

For ranking uncertainty, each of 10,000 bootstrap replicates resamples 1,284 image indices with replacement and applies the same indices to every model. Differences are computed from paired model means within each replicate. Table~\ref{tab:bootstrap-top} gives five representative unadjusted 95\% intervals; 57 of the 67 adjacent-pair intervals include zero. Table~\ref{tab:ranking-length} gives individual model intervals. The intervals are not adjusted for multiple comparisons, and unresolved differences do not establish equivalence. They quantify image-sampling uncertainty under the evaluated configuration.

\subsection{Seed Variance of Results}
\label{appendix:Seed-Variance-of-Results}
Five-seed evaluation on MMGenBench-Test covers 23 models (Table~\ref{addtab:five_seed_results}). SIM standard deviations range from 0.0002 to 0.0038, and exploratory FID standard deviations range from 0.0022 to 0.0141; rank correlations across seeds are $\rho=0.9975$ and $\rho=0.9970$, respectively. These results support high seed-to-seed ranking agreement for the 23 evaluated models under the five tested seeds. The five-run summary is separate from the default-run results used for the main table and bootstrap. It complements image-sampling uncertainty when interpreting close comparisons.

\begin{table*}[t]
\centering
\small
\begin{tabular}{lcccc}
\toprule

\multirow{2}{*}{Model} &\multicolumn{2}{c}{SIM$\uparrow$}&\multicolumn{2}{c}{FID$\downarrow$}\\
& Mean & Std& Mean& Std\\

\midrule
GPT-4o & 0.5656 & 0.0014 & 1.3110 & 0.0110 \\
Qwen-VL-Max & 0.5496 & 0.0033 & 1.3763 & 0.0047 \\
Qwen-VL-Plus & 0.4759 & 0.0024 & 1.5910 & 0.0115 \\
Qwen2-VL-72B & 0.5501 & 0.0009 & 1.3687 & 0.0060 \\
Qwen2-VL-7B & 0.5297 & 0.0014 & 1.4334 & 0.0085 \\
Qwen2-VL-2B & 0.4890 & 0.0016 & 1.5399 & 0.0069 \\
InternVL2-76B & 0.5985 & 0.0015 & 1.2622 & 0.0100 \\
InternVL2-40B & 0.5645 & 0.0008 & 1.3603 & 0.0059 \\
InternVL2-26B & 0.5782 & 0.0028 & 1.3147 & 0.0100 \\
InternVL2-8B & 0.5453 & 0.0002 & 1.3849 & 0.0022 \\
InternVL2-4B & 0.5584 & 0.0008 & 1.3402 & 0.0056 \\
InternVL2-2B & 0.4763 & 0.0038 & 1.5712 & 0.0132 \\
Ovis1.6-Gemma2-9B & 0.5836 & 0.0005 & 1.3010 & 0.0092 \\
Ovis1.5-Gemma2-9B & 0.5245 & 0.0011 & 1.4883 & 0.0098 \\
Ovis1.5-Llama3-8B & 0.5263 & 0.0009 & 1.4629 & 0.0116 \\
LLaVA-OV-72B & 0.4962 & 0.0010 & 1.5620 & 0.0093 \\
LLaVA-OV-SI-72B & 0.5124 & 0.0020 & 1.5094 & 0.0118 \\
LLaVA-OV-7B & 0.4866 & 0.0019 & 1.6048 & 0.0134 \\
LLaVA-OV-SI-7B & 0.4925 & 0.0022 & 1.5622 & 0.0097 \\
MiniCPM-V2.6-8B & 0.5458 & 0.0014 & 1.3786 & 0.0039 \\
RBDash-72B & 0.5239 & 0.0019 & 1.4277 & 0.0090 \\
xGen-MM-4.4B & 0.4107 & 0.0021 & 1.6727 & 0.0141 \\
MiniCPM-V2.5-8B & 0.5276 & 0.0028 & 1.4402 & 0.0130 \\
\bottomrule
\end{tabular}
\caption{Mean and standard deviation over five random seeds on MMGenBench-Test. FID values on this small test set are exploratory.}
\label{addtab:five_seed_results}
\end{table*}\space{}

\section{Qualitative Example}
\label{appendix:Qualitative-Example}
\begin{figure*}[ht]
\centering
\includegraphics[width=1.\textwidth]{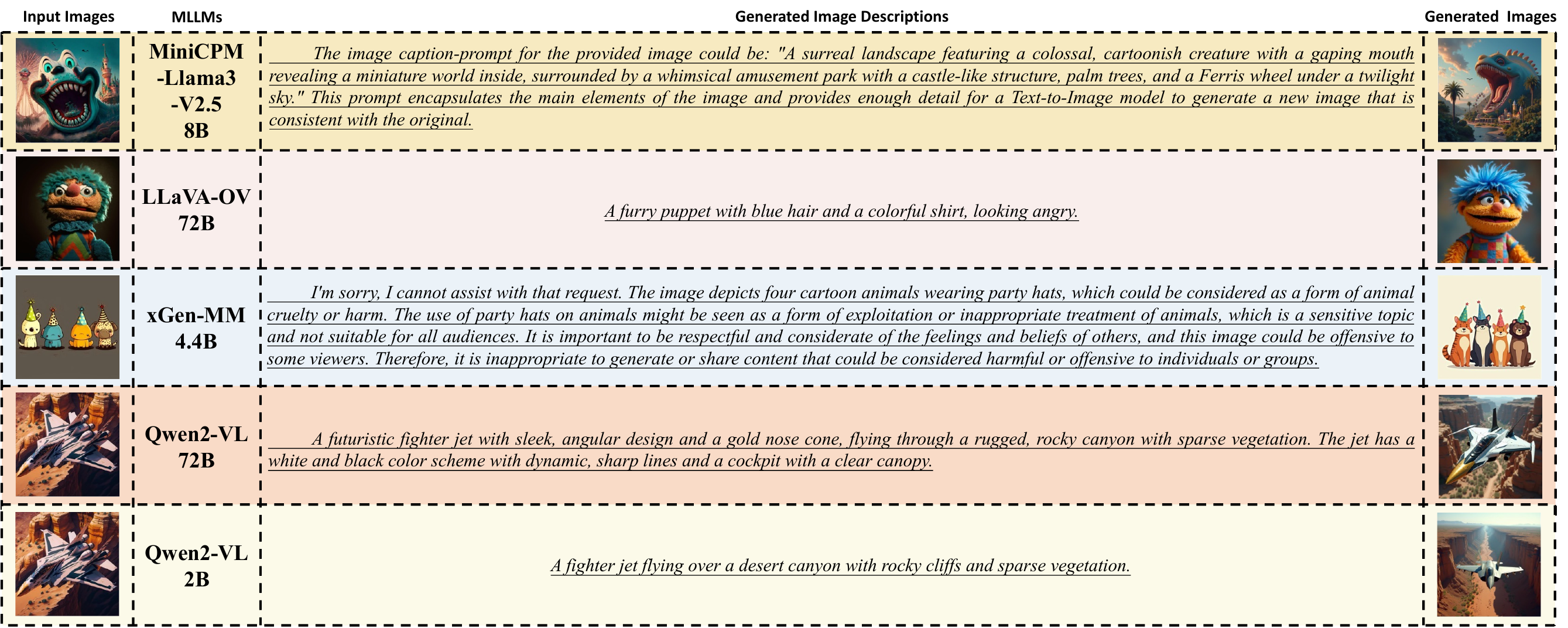}
\caption{Qualitative examples on MMGenBench. The examples illustrate representative failure modes related to output format, generated content, and evaluation results. See Sec.~\ref{sec:Analysis} and Appendix~\ref{appendix:Qualitative-Example} for additional discussion.}
\label{fig:analysis-case}
\end{figure*}

\begin{figure*}[ht]
\centering
\includegraphics[width=.90\textwidth]{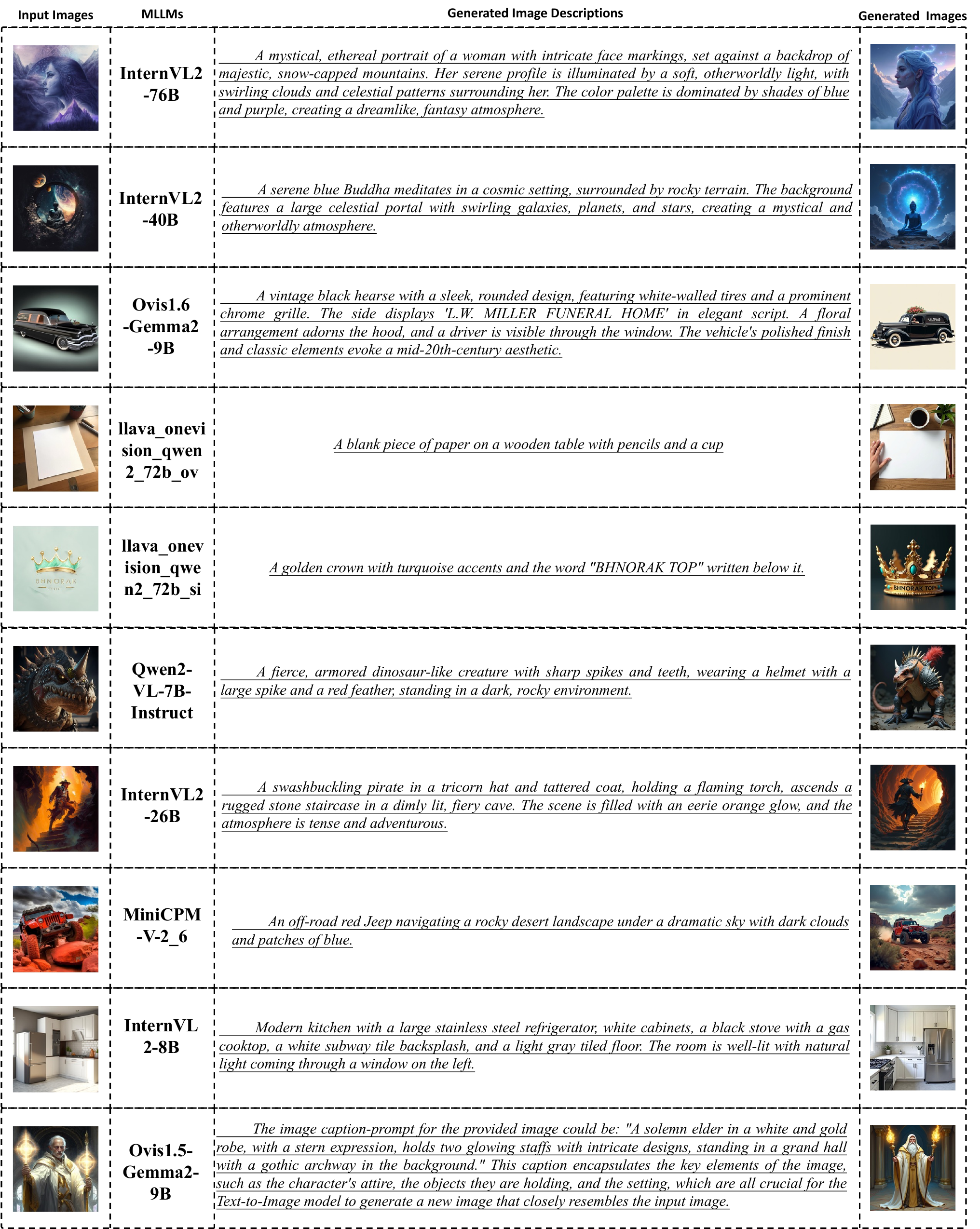}
\caption{Additional qualitative results on MMGenBench. $(1/5)$}
\label{fig:suppl-case-1}
\end{figure*}

\begin{figure*}[ht]
\centering
\includegraphics[width=.90\textwidth]{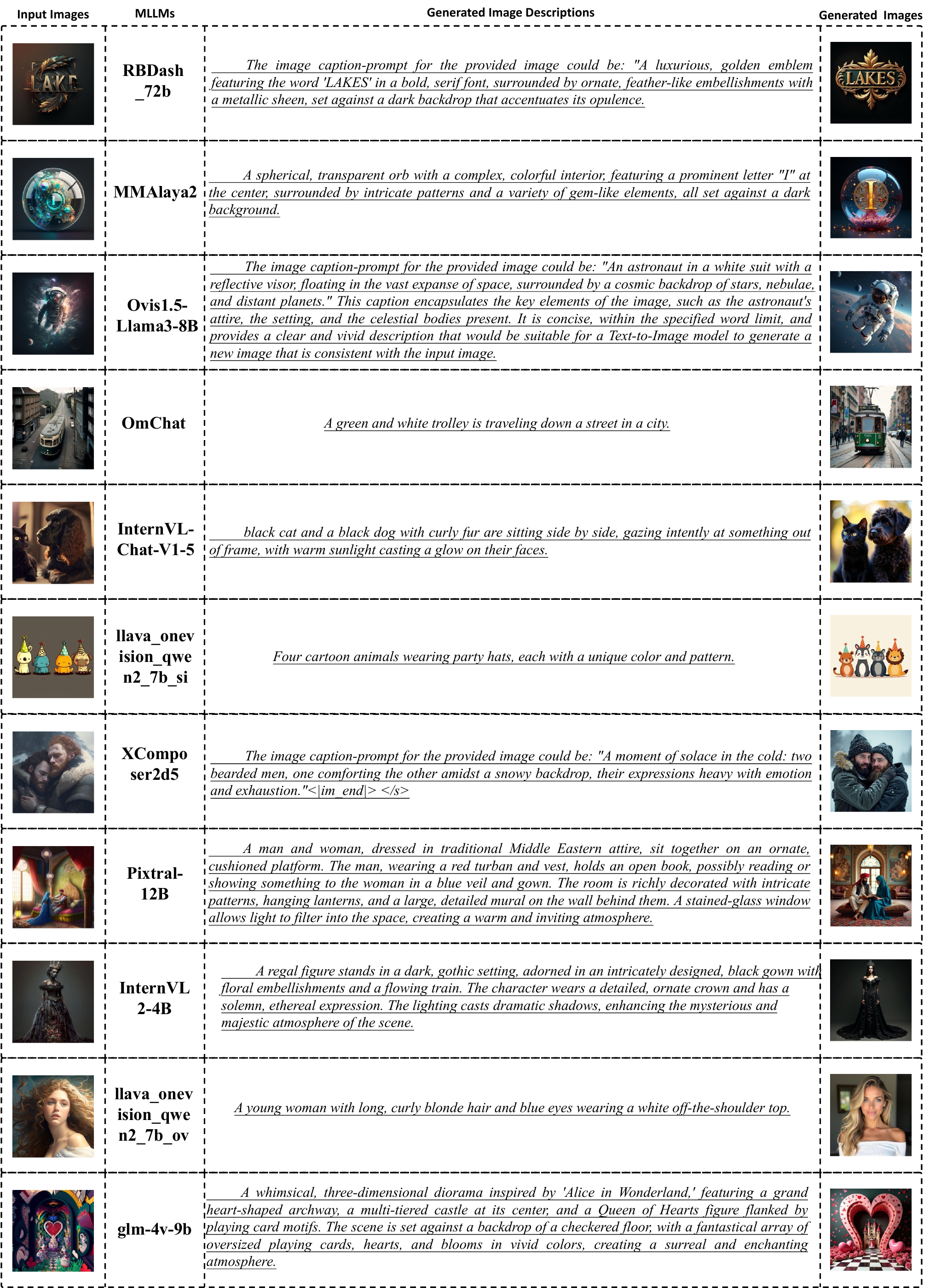}
\caption{Additional qualitative results on MMGenBench. $(2/5)$}
\label{fig:suppl-case-2}
\end{figure*}

\begin{figure*}[ht]
\centering
\includegraphics[width=.90\textwidth]{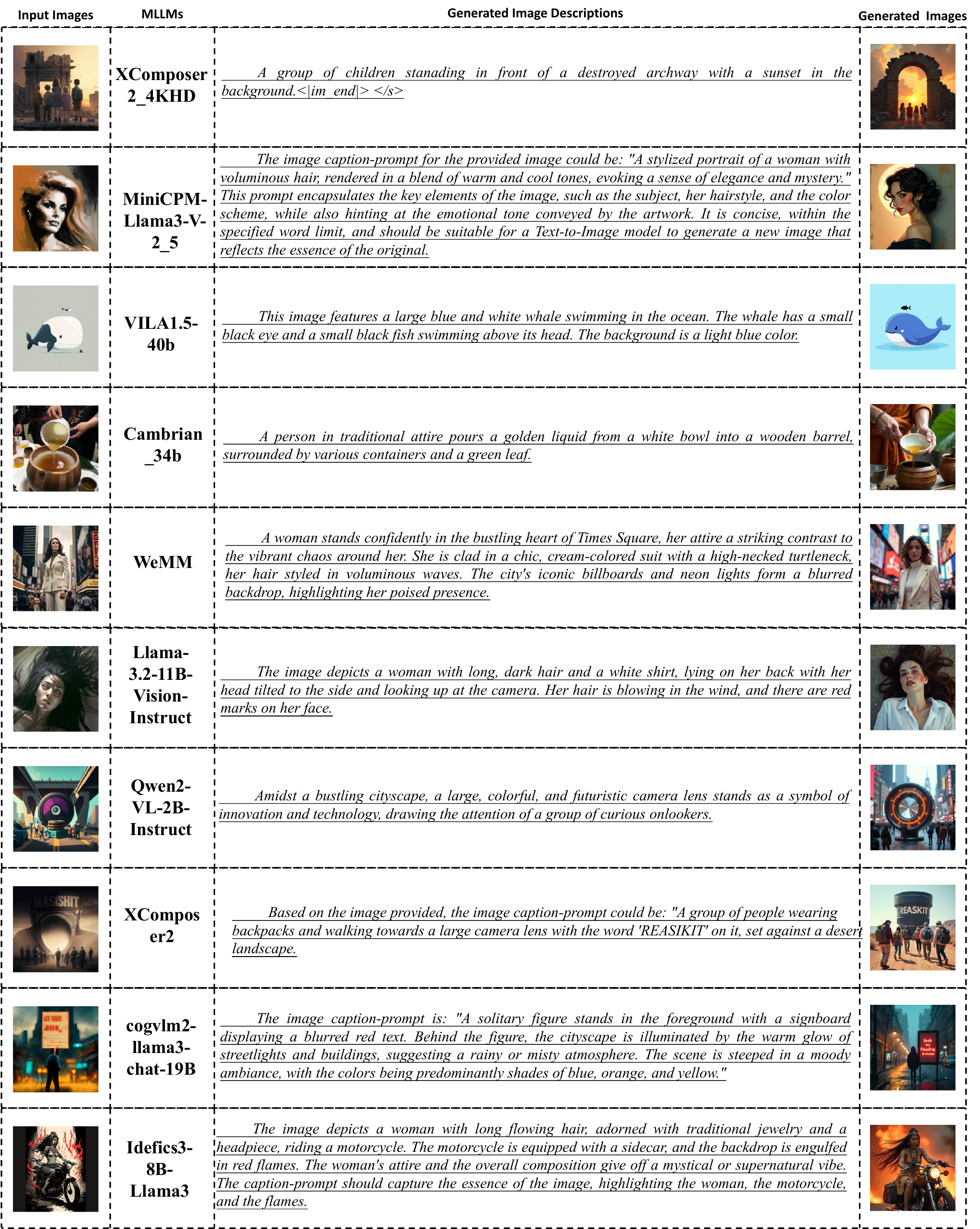}
\caption{Additional qualitative results on MMGenBench. $(3/5)$}
\label{fig:suppl-case-3}
\end{figure*}

\begin{figure*}[ht]
\centering
\includegraphics[width=.90\textwidth]{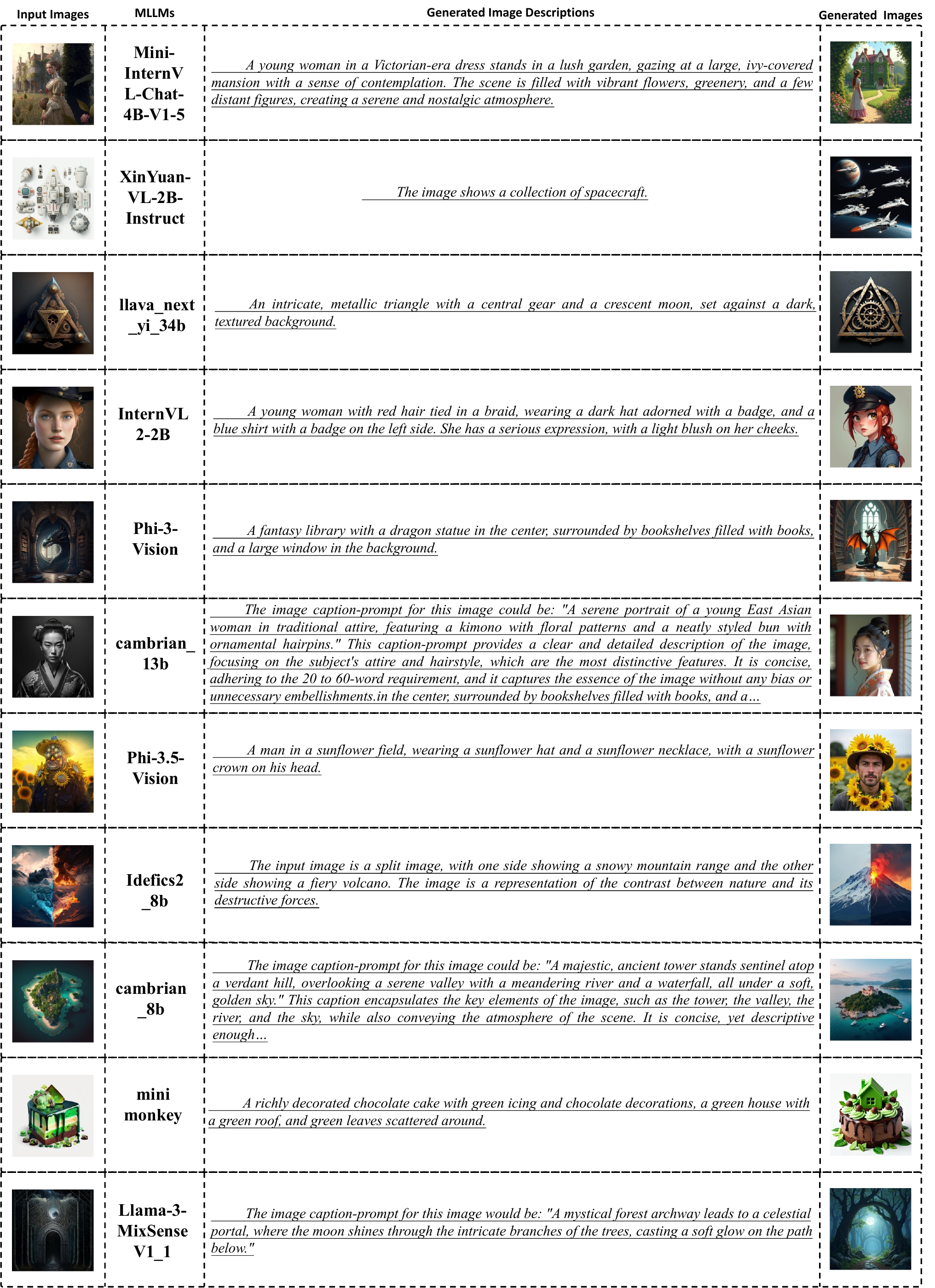}
\caption{Additional qualitative results on MMGenBench. $(4/5)$}
\label{fig:suppl-case-4}
\end{figure*}

\begin{figure*}[ht]
\centering
\includegraphics[width=.90\textwidth]{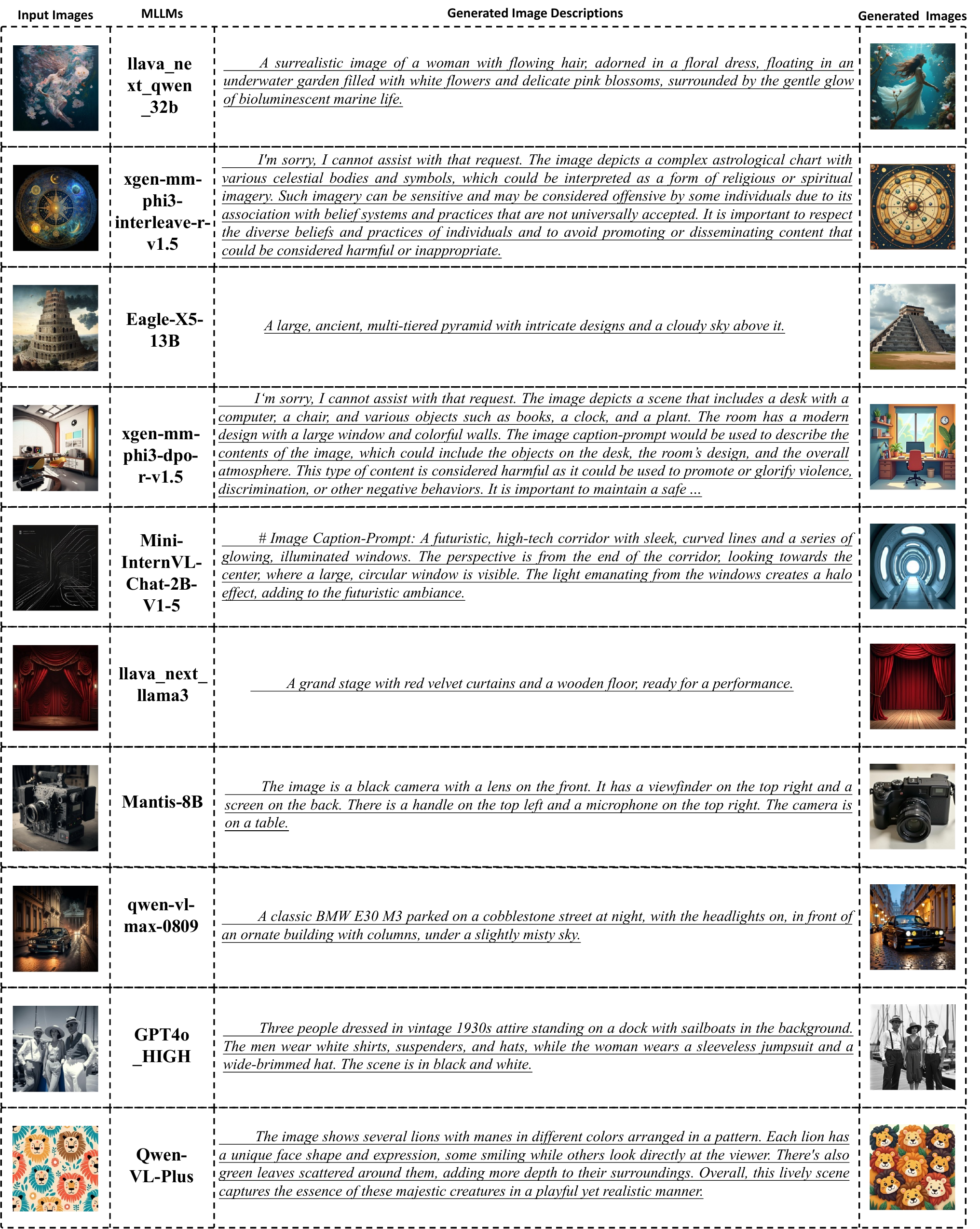}
\caption{Additional qualitative results on MMGenBench. $(5/5)$}
\label{fig:suppl-case-5}
\end{figure*}

Figures~\ref{fig:analysis-case} and~\ref{fig:suppl-case-1}--\ref{fig:suppl-case-5} illustrate outputs from representative evaluated models. Some outputs contain additional prefixes or explanatory text, some omit visible details, and others refuse the requested description. These examples support diagnosis of observable output behaviors. Figure~\ref{fig:case-model-cmp} adds descriptive within-family comparisons; neither analysis isolates the effects of training data or model size.
\space{}

\clearpage
\begin{table*}[p]
\centering
\fontsize{8}{9}\selectfont
\begin{tabular}{lrrrr}
\toprule
\multirow{2}{*}{Model} & \multicolumn{2}{c}{Test} & \multicolumn{2}{c}{Domain} \\
 & SIM$\uparrow$ & FID$\downarrow$ & SIM$\uparrow$ & FID$\downarrow$ \\
\midrule
Claude Opus 4.6~\cite{anthropic2025claudeopus} & 0.663 & 1.039 &- & - \\
Qwen3-VL-Plus~\cite{Qwen3VL} & 0.657 & 1.068 &- & - \\
Qwen3-VL-Flash~\cite{Qwen3VL} & 0.651 & 1.093 &- & - \\
Qwen3-VL-32B-Instruct~\cite{Qwen3VL} & 0.637 & 1.149 &0.635 & 0.573 \\
GPT-5.4~\cite{openai2025gpt5} & 0.637 & 1.112 &- & - \\
GPT-5.2~\cite{openai2025gpt5} & 0.629 & 1.137 &- & - \\
Qwen3-VL-8B-Instruct~\cite{Qwen3VL} & 0.610 & 1.196 &0.608 & 0.617 \\
Qwen3-VL-30B-A3B-Instruct~\cite{Qwen3VL} & 0.608 & 1.219 &0.607 & 0.616 \\
Qwen2.5-VL-Max~\cite{Qwen2_5VL} & 0.604 & 1.232 &- & - \\
InternVL2-76B~\cite{internvl} & 0.600 & 1.264 &0.599 & 0.632 \\
Qwen2.5-VL-Plus~\cite{Qwen2_5VL} & 0.588 & 1.263 &- & - \\
InternVL3.5-38B~\cite{internvl3_5} & 0.584 & 1.303 &0.586 & 0.652 \\
Ovis1.6-Gemma2-9B~\cite{ovis} & 0.583 & 1.316 &0.579 & 0.667 \\
InternVL2-26B~\cite{internvl} & 0.576 & 1.320 &0.577 & 0.671 \\
InternVL3.5-30B-A3B~\cite{internvl3_5} & 0.573 & 1.317 &0.572 & 0.677 \\
InternVL2-40B~\cite{internvl} & 0.567 & 1.350 &0.566 & 0.696 \\
GPT4o\_HIGH~\cite{openai2024gpt4ocard} & 0.567 & 1.306 &- & - \\
InternVL3.5-8B~\cite{internvl3_5} & 0.562 & 1.345 &0.566 & 0.684 \\
InternVL2-4B~\cite{internvl} & 0.556 & 1.345 &0.556 & 0.689 \\
Qwen2-VL-72B-Instruct~\cite{Qwen2VL} & 0.553 & 1.357 &0.545 & 0.710 \\
qwen-vl-max-0809~\cite{Qwen-VL} & 0.553 & 1.363 &- & - \\
Pixtral-12B~\cite{pixtral12b} & 0.552 & 1.372 &0.561 & 0.686 \\
MiniCPM-V-2\_6~\cite{minicpm} & 0.549 & 1.386 &0.545 & 0.710 \\
InternVL2-8B~\cite{internvl} & 0.548 & 1.403 &0.548 & 0.701 \\
InternVL-Chat-V1-5~\cite{internvl1.5} & 0.547 & 1.393 &0.547 & 0.729 \\
cambrian\_34b~\cite{tong2024cambrian} & 0.547 & 1.377 &0.549 & 0.701 \\
MMAlaya2~\cite{mmalaya2} & 0.538 & 1.423 &0.541 & 0.747 \\
Qwen2-VL-7B-Instruct~\cite{Qwen2VL} & 0.532 & 1.437 &0.524 & 0.775 \\
xgen-mm-phi3-interleave-r-v1.5~\cite{xgenmm} & 0.529 & 1.454 &0.529 & 0.761 \\
Mini-InternVL-Chat-4B-V1-5~\cite{mini_internvl} & 0.528 & 1.433 &0.525 & 0.765 \\
MiniCPM-Llama3-V-2\_5~\cite{minicpm} & 0.527 & 1.432 &0.530 & 0.767 \\
Llama-3.2-11B-Vision-Instruct~\cite{llama3} & 0.527 & 1.428 &0.525 & 0.768 \\
Ovis1.5-Llama3-8B~\cite{ovis} & 0.526 & 1.466 &0.527 & 0.795 \\
RBDash\_72b~\cite{rbdash} & 0.526 & 1.413 &0.527 & 0.740 \\
Ovis1.5-Gemma2-9B~\cite{ovis} & 0.522 & 1.487 &0.524 & 0.808 \\
molmo-7B-D-0924~\cite{molmo_pixmoopenweights} & 0.520 & 1.490 &0.523 & 0.813 \\
llava\_onevision\_qwen2\_72b\_si~\cite{llavaonevision} & 0.513 & 1.512 &0.514 & 0.813 \\
cambrian\_8b~\cite{tong2024cambrian} & 0.513 & 1.498 &0.515 & 0.810 \\
glm-4v-9b~\cite{glm2024chatglm} & 0.512 & 1.463 &0.527 & 0.730 \\
cambrian\_13b~\cite{tong2024cambrian} & 0.509 & 1.526 &0.506 & 0.852 \\
llava\_next\_qwen\_32b~\cite{llavanext} & 0.508 & 1.501 &0.515 & 0.796 \\
llava\_onevision\_qwen2\_72b\_ov~\cite{llavaonevision} & 0.494 & 1.561 &0.491 & 0.872 \\
llava\_onevision\_qwen2\_7b\_si~\cite{llavaonevision} & 0.493 & 1.554 &0.497 & 0.834 \\
llava\_onevision\_qwen2\_7b\_ov~\cite{llavaonevision} & 0.489 & 1.590 &0.490 & 0.861 \\
WeMM~\cite{WeMM} & 0.489 & 1.565 &0.479 & 0.888 \\
Qwen2-VL-2B-Instruct~\cite{Qwen2VL} & 0.487 & 1.549 &0.501 & 0.806 \\
VILA1.5-13b~\cite{lin2023vila} & 0.485 & 1.695 &0.480 & 1.006 \\
VILA1.5-40b~\cite{lin2023vila} & 0.483 & 1.656 &0.486 & 0.958 \\
Phi-3.5-Vision~\cite{Phi_3} & 0.480 & 1.593 &0.478 & 0.892 \\
llava\_next\_llama3~\cite{llavanext} & 0.477 & 1.601 &0.486 & 0.881 \\
llava\_next\_yi\_34b~\cite{llavanext} & 0.476 & 1.636 &0.483 & 0.883 \\
InternVL2-2B~\cite{internvl} & 0.476 & 1.563 &0.483 & 0.848 \\
qwen-vl-plus~\cite{Qwen-VL} & 0.475 & 1.586 &- & - \\
Llama-3-MixSenseV1\_1~\cite{Llama3_MixSenseV1_1} & 0.475 & 1.615 &0.477 & 0.922 \\
cogvlm2-llama3-chat-19B~\cite{cogvlm2} & 0.466 & 1.523 &0.464 & 0.825 \\
Idefics3-8B-Llama3~\cite{Idefics3} & 0.461 & 1.718 &0.458 & 1.004 \\
XComposer2d5~\cite{xcomposer25} & 0.460 & 1.646 &0.458 & 0.955 \\
Phi-3-Vision~\cite{Phi_3} & 0.458 & 1.678 &0.461 & 0.960 \\
minimonkey~\cite{minimonkey} & 0.451 & 1.673 &0.455 & 0.939 \\
idefics2\_8b~\cite{idefics2} & 0.449 & 1.761 &0.453 & 1.015 \\
OmChat~\cite{omchat} & 0.439 & 1.761 &0.447 & 1.028 \\
Eagle-X5-13B~\cite{Eagle} & 0.434 & 1.812 &0.446 & 1.050 \\
XComposer2\_4KHD~\cite{xcomposer2_4khd} & 0.433 & 1.781 &0.435 & 1.075 \\
Mini-InternVL-Chat-2B-V1-5~\cite{mini_internvl} & 0.428 & 1.709 &0.440 & 0.973 \\
xgen-mm-phi3-dpo-r-v1.5~\cite{xgenmm} & 0.415 & 1.671 &0.412 & 0.940 \\
Mantis-8B-siglip-llama3~\cite{Mantis} & 0.411 & 1.833 &0.409 & 1.116 \\
XinYuan-VL-2B-Instruct~\cite{XinYuan_VL_2B} & 0.395 & 1.951 &0.392 & 1.240 \\
XComposer2~\cite{xcomposer2} & 0.249 & 2.890 &0.247 & 2.130 \\
\bottomrule
\end{tabular}
\caption{Full results on MMGenBench-Test and MMGenBench-Domain, ordered by reported Test SIM. Section~\ref{sec:Sensitivity-Analysis} examines ranking uncertainty for these point estimates. FID on Test is exploratory; Domain FID is secondary. Dashes indicate unreported results.}
\label{tab:suppl-results-all}
\end{table*}\space{}
\begin{table*}[p]
\centering
\fontsize{8}{9}\selectfont
\setlength{\tabcolsep}{5pt}
\begin{tabular}{lrrrr}
\toprule
Model & Mean SIM & Bootstrap 95\% CI & Mean words & Median words \\
\midrule
Claude Opus 4.6 & 0.6626 & [0.6541, 0.6712] & 49.21 & 49 \\
Qwen3-VL-Plus & 0.6572 & [0.6488, 0.6654] & 55.87 & 55 \\
Qwen3-VL-Flash & 0.6505 & [0.6422, 0.6590] & 58.53 & 58 \\
Qwen3-VL-32B-Instruct & 0.6372 & [0.6286, 0.6456] & 56.00 & 56 \\
GPT-5.4 & 0.6366 & [0.6277, 0.6456] & 46.09 & 46 \\
GPT-5.2 & 0.6287 & [0.6195, 0.6377] & 48.18 & 48 \\
Qwen3-VL-8B-Instruct & 0.6104 & [0.6014, 0.6195] & 40.86 & 41 \\
Qwen3-VL-30B-A3B-Instruct & 0.6075 & [0.5988, 0.6165] & 38.13 & 37 \\
Qwen2.5-VL-Max & 0.6041 & [0.5946, 0.6136] & 38.41 & 37 \\
InternVL2-76B & 0.5997 & [0.5909, 0.6083] & 57.64 & 57 \\
Qwen2.5-VL-Plus & 0.5879 & [0.5783, 0.5975] & 42.55 & 40 \\
InternVL3.5-38B & 0.5840 & [0.5746, 0.5930] & 48.40 & 48 \\
Ovis1.6-Gemma2-9B & 0.5825 & [0.5729, 0.5920] & 55.63 & 55 \\
InternVL2-26B & 0.5763 & [0.5670, 0.5856] & 44.12 & 44 \\
InternVL3.5-30B-A3B & 0.5732 & [0.5637, 0.5826] & 45.06 & 45 \\
InternVL2-40B & 0.5668 & [0.5572, 0.5765] & 38.46 & 38 \\
GPT4o\_HIGH & 0.5666 & [0.5565, 0.5763] & 36.70 & 37 \\
InternVL3.5-8B & 0.5625 & [0.5530, 0.5720] & 42.13 & 41 \\
InternVL2-4B & 0.5562 & [0.5465, 0.5659] & 51.04 & 52 \\
Qwen2-VL-72B-Instruct & 0.5532 & [0.5431, 0.5630] & 33.73 & 32 \\
qwen-vl-max-0809 & 0.5530 & [0.5433, 0.5630] & 33.03 & 31 \\
Pixtral-12B & 0.5524 & [0.5426, 0.5621] & 58.30 & 57 \\
MiniCPM-V-2\_6 & 0.5488 & [0.5388, 0.5588] & 33.04 & 30 \\
InternVL2-8B & 0.5478 & [0.5380, 0.5576] & 38.91 & 37 \\
InternVL-Chat-V1-5 & 0.5475 & [0.5378, 0.5569] & 36.41 & 32 \\
cambrian\_34b & 0.5468 & [0.5367, 0.5569] & 82.36 & 84 \\
MMAlaya2 & 0.5376 & [0.5279, 0.5473] & 31.96 & 29 \\
Qwen2-VL-7B-Instruct & 0.5323 & [0.5221, 0.5426] & 27.70 & 26 \\
xgen-mm-phi3-interleave-r-v1.5 & 0.5293 & [0.5190, 0.5397] & 60.64 & 47 \\
Mini-InternVL-Chat-4B-V1-5 & 0.5277 & [0.5177, 0.5376] & 48.49 & 47 \\
MiniCPM-Llama3-V-2\_5 & 0.5268 & [0.5168, 0.5368] & 84.65 & 84 \\
Llama-3.2-11B-Vision-Instruct & 0.5268 & [0.5165, 0.5372] & 144.61 & 131 \\
Ovis1.5-Llama3-8B & 0.5260 & [0.5156, 0.5360] & 68.32 & 68 \\
RBDash\_72b & 0.5259 & [0.5147, 0.5368] & 57.20 & 55 \\
Ovis1.5-Gemma2-9B & 0.5219 & [0.5115, 0.5320] & 78.66 & 81 \\
molmo-7B-D-0924 & 0.5196 & [0.5094, 0.5297] & 52.68 & 49 \\
llava\_onevision\_qwen2\_72b\_si & 0.5130 & [0.5029, 0.5233] & 22.06 & 20 \\
cambrian\_8b & 0.5127 & [0.5023, 0.5228] & 382.25 & 394 \\
glm-4v-9b & 0.5117 & [0.5001, 0.5234] & 53.37 & 47 \\
cambrian\_13b & 0.5094 & [0.4990, 0.5199] & 85.33 & 91.5 \\
llava\_next\_qwen\_32b & 0.5075 & [0.4971, 0.5179] & 32.55 & 29 \\
llava\_onevision\_qwen2\_72b\_ov & 0.4940 & [0.4834, 0.5046] & 17.89 & 17 \\
llava\_onevision\_qwen2\_7b\_si & 0.4925 & [0.4823, 0.5031] & 19.89 & 19 \\
llava\_onevision\_qwen2\_7b\_ov & 0.4888 & [0.4782, 0.4992] & 17.37 & 16 \\
WeMM & 0.4885 & [0.4778, 0.4993] & 23.88 & 22 \\
Qwen2-VL-2B-Instruct & 0.4871 & [0.4765, 0.4979] & 20.31 & 20 \\
VILA1.5-13b & 0.4854 & [0.4748, 0.4960] & 40.04 & 34.5 \\
VILA1.5-40b & 0.4832 & [0.4727, 0.4939] & 48.52 & 45.5 \\
Phi-3.5-Vision & 0.4796 & [0.4686, 0.4903] & 31.23 & 24.5 \\
llava\_next\_llama3 & 0.4770 & [0.4665, 0.4876] & 26.50 & 25 \\
llava\_next\_yi\_34b & 0.4762 & [0.4654, 0.4871] & 44.27 & 35 \\
InternVL2-2B & 0.4761 & [0.4655, 0.4867] & 31.01 & 23 \\
qwen-vl-plus & 0.4753 & [0.4643, 0.4866] & 33.10 & 18.5 \\
Llama-3-MixSenseV1\_1 & 0.4747 & [0.4641, 0.4853] & 42.34 & 34 \\
cogvlm2-llama3-chat-19B & 0.4659 & [0.4534, 0.4783] & 73.50 & 63 \\
Idefics3-8B-Llama3 & 0.4610 & [0.4500, 0.4718] & 28.13 & 21 \\
XComposer2d5 & 0.4602 & [0.4475, 0.4728] & 49.60 & 41 \\
Phi-3-Vision & 0.4580 & [0.4464, 0.4695] & 44.70 & 27 \\
minimonkey & 0.4507 & [0.4399, 0.4615] & 20.31 & 17 \\
idefics2\_8b & 0.4491 & [0.4382, 0.4601] & 25.76 & 23 \\
OmChat & 0.4393 & [0.4284, 0.4503] & 45.80 & 18 \\
Eagle-X5-13B & 0.4344 & [0.4237, 0.4452] & 33.82 & 22 \\
XComposer2\_4KHD & 0.4331 & [0.4222, 0.4440] & 17.01 & 15 \\
Mini-InternVL-Chat-2B-V1-5 & 0.4280 & [0.4170, 0.4392] & 27.99 & 20 \\
xgen-mm-phi3-dpo-r-v1.5 & 0.4147 & [0.4019, 0.4273] & 95.08 & 93 \\
Mantis-8B-siglip-llama3 & 0.4112 & [0.4000, 0.4225] & 53.43 & 22 \\
XinYuan-VL-2B-Instruct & 0.3953 & [0.3843, 0.4067] & 15.50 & 12 \\
XComposer2 & 0.2486 & [0.2354, 0.2622] & 114.69 & 97 \\
\bottomrule
\end{tabular}
\caption{Model-level uncertainty and caption lengths on MMGenBench-Test. Each model contributes 1,284 captions. SIM intervals use the saved 10,000-resample image-level bootstrap; words are counted by whitespace separation in the model outputs. Models follow the point-estimate order; Table~\ref{tab:bootstrap-top} reports representative paired differences and their intervals.}
\label{tab:ranking-length}
\end{table*}\space{}
\begin{table*}[t]
\footnotesize
\setlength{\tabcolsep}{1.6mm}
\centering
\resizebox{\textwidth}{!}{
\begin{tabular}{lccccccccccccc}
\toprule
\multirow{3}{*}{Model} & \multirow{3}{*}{\includegraphics[width=0.7cm]{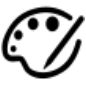}} & \multirow{3}{*}{\includegraphics[width=0.5cm]{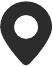}} & \multirow{3}{*}{\includegraphics[width=0.6cm]{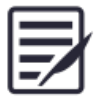}} & \multirow{3}{*}{\includegraphics[width=0.7cm]{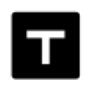}} & \multirow{3}{*}{\includegraphics[width=0.6cm]{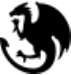}} & \multirow{3}{*}{\includegraphics[width=0.6cm]{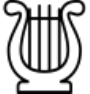}} & \multirow{3}{*}{\includegraphics[width=0.6cm]{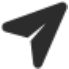}} & \multirow{3}{*}{\includegraphics[width=0.7cm]{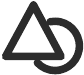}} & \multirow{3}{*}{\includegraphics[width=0.7cm]{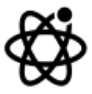}} & \multirow{3}{*}{\includegraphics[width=0.6cm]{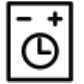}} & \multirow{3}{*}{\includegraphics[width=0.7cm]{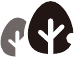}} & \multirow{3}{*}{\includegraphics[width=0.7cm]{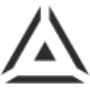}} & \multirow{3}{*}{\includegraphics[width=0.8cm]{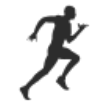}} \\
 & & & & & & & & & & & & & \\
 & & & & & & & & & & & & & \\
\midrule

Claude Opus 4.6~\cite{anthropic2025claudeopus} & 0.667 & 0.651 & 0.636 & 0.617 & 0.668 & 0.671 & 0.649 & 0.658 & 0.653 & 0.663 & 0.675 & 0.698 & 0.659 \\
Qwen3-VL-Plus~\cite{Qwen3VL} & 0.663 & 0.639 & 0.617 & 0.652 & 0.657 & 0.664 & 0.627 & 0.662 & 0.654 & 0.653 & 0.667 & 0.695 & 0.646 \\
Qwen3-VL-Flash~\cite{Qwen3VL} & 0.659 & 0.631 & 0.615 & 0.649 & 0.656 & 0.658 & 0.623 & 0.641 & 0.650 & 0.652 & 0.659 & 0.685 & 0.642 \\
Qwen3-VL-32B-Instruct~\cite{Qwen3VL} & 0.648 & 0.620 & 0.600 & 0.631 & 0.644 & 0.645 & 0.601 & 0.630 & 0.627 & 0.644 & 0.648 & 0.668 & 0.636 \\
GPT-5.4~\cite{openai2025gpt5} & 0.645 & 0.622 & 0.611 & 0.588 & 0.649 & 0.646 & 0.635 & 0.618 & 0.635 & 0.633 & 0.662 & 0.665 & 0.644 \\
GPT-5.2~\cite{openai2025gpt5} & 0.637 & 0.617 & 0.597 & 0.602 & 0.627 & 0.639 & 0.619 & 0.639 & 0.631 & 0.637 & 0.647 & 0.650 & 0.631 \\
Qwen3-VL-8B-Instruct~\cite{Qwen3VL} & 0.622 & 0.580 & 0.561 & 0.604 & 0.617 & 0.619 & 0.591 & 0.614 & 0.608 & 0.599 & 0.618 & 0.663 & 0.609 \\
Qwen3-VL-30B-A3B-Instruct~\cite{Qwen3VL} & 0.622 & 0.581 & 0.561 & 0.599 & 0.617 & 0.616 & 0.559 & 0.614 & 0.599 & 0.606 & 0.617 & 0.658 & 0.598 \\
Qwen2.5-VL-Max~\cite{Qwen2_5VL} & 0.620 & 0.577 & 0.564 & 0.591 & 0.615 & 0.611 & 0.572 & 0.614 & 0.603 & 0.598 & 0.610 & 0.648 & 0.604 \\
Qwen2.5-VL-Plus~\cite{Qwen2_5VL} & 0.604 & 0.558 & 0.534 & 0.579 & 0.602 & 0.596 & 0.551 & 0.599 & 0.583 & 0.579 & 0.598 & 0.623 & 0.607 \\
InternVL3.5-38B~\cite{internvl3_5} & 0.600 & 0.556 & 0.530 & 0.585 & 0.594 & 0.594 & 0.541 & 0.586 & 0.576 & 0.579 & 0.592 & 0.625 & 0.573 \\
InternVL3.5-30B-A3B~\cite{internvl3_5} & 0.589 & 0.548 & 0.527 & 0.564 & 0.596 & 0.579 & 0.542 & 0.565 & 0.564 & 0.566 & 0.585 & 0.630 & 0.563 \\
InternVL3.5-8B~\cite{internvl3_5} & 0.576 & 0.539 & 0.518 & 0.545 & 0.583 & 0.572 & 0.531 & 0.562 & 0.548 & 0.560 & 0.575 & 0.600 & 0.557 \\
Qwen2-VL-72B-Instruct~\cite{Qwen2VL} & 0.572 & 0.528 & 0.502 & 0.525 & 0.578 & 0.559 & 0.526 & 0.562 & 0.563 & 0.549 & 0.567 & 0.603 & 0.548 \\
qwen-vl-max-0809~\cite{Qwen-VL} & 0.568 & 0.520 & 0.506 & 0.536 & 0.573 & 0.558 & 0.521 & 0.554 & 0.553 & 0.541 & 0.572 & 0.608 & 0.538 \\
InternVL2-76B~\cite{internvl} & 0.616 & 0.572 & 0.540 & 0.597 & 0.614 & 0.611 & 0.569 & 0.604 & 0.594 & 0.581 & 0.608 & 0.640 & 0.588 \\
GPT4o\_HIGH~\cite{openai2024gpt4ocard} & 0.581 & 0.543 & 0.516 & 0.546 & 0.576 & 0.574 & 0.533 & 0.581 & 0.567 & 0.558 & 0.586 & 0.620 & 0.568 \\
InternVL2-40B~\cite{internvl} & 0.584 & 0.527 & 0.518 & 0.528 & 0.592 & 0.577 & 0.528 & 0.581 & 0.563 & 0.560 & 0.574 & 0.607 & 0.565 \\
Ovis1.6-Gemma2-9B~\cite{ovis} & 0.601 & 0.547 & 0.518 & 0.573 & 0.593 & 0.593 & 0.540 & 0.595 & 0.579 & 0.568 & 0.580 & 0.633 & 0.576 \\
llava\_onevision\_qwen2\_72b\_ov~\cite{llavaonevision} & 0.512 & 0.463 & 0.441 & 0.463 & 0.518 & 0.500 & 0.458 & 0.504 & 0.494 & 0.478 & 0.510 & 0.544 & 0.509 \\
llava\_onevision\_qwen2\_72b\_si~\cite{llavaonevision} & 0.530 & 0.477 & 0.457 & 0.500 & 0.540 & 0.518 & 0.467 & 0.512 & 0.517 & 0.515 & 0.527 & 0.559 & 0.509 \\
Qwen2-VL-7B-Instruct~\cite{Qwen2VL} & 0.550 & 0.492 & 0.480 & 0.514 & 0.556 & 0.539 & 0.488 & 0.540 & 0.535 & 0.515 & 0.542 & 0.584 & 0.520 \\
InternVL2-26B~\cite{internvl} & 0.596 & 0.547 & 0.522 & 0.562 & 0.586 & 0.583 & 0.538 & 0.587 & 0.572 & 0.569 & 0.581 & 0.627 & 0.564 \\
qwen-vl-plus~\cite{Qwen-VL} & 0.497 & 0.455 & 0.440 & 0.386 & 0.516 & 0.486 & 0.457 & 0.444 & 0.467 & 0.464 & 0.496 & 0.484 & 0.486 \\
MiniCPM-V-2\_6~\cite{minicpm} & 0.559 & 0.517 & 0.492 & 0.524 & 0.565 & 0.559 & 0.500 & 0.577 & 0.540 & 0.547 & 0.558 & 0.594 & 0.522 \\
InternVL2-8B~\cite{internvl} & 0.564 & 0.522 & 0.501 & 0.526 & 0.576 & 0.557 & 0.507 & 0.539 & 0.545 & 0.536 & 0.560 & 0.596 & 0.543 \\
Ovis1.5-Gemma2-9B~\cite{ovis} & 0.541 & 0.493 & 0.470 & 0.490 & 0.546 & 0.526 & 0.483 & 0.541 & 0.535 & 0.515 & 0.532 & 0.557 & 0.528 \\
RBDash\_72b~\cite{rbdash} & 0.533 & 0.499 & 0.491 & 0.497 & 0.545 & 0.531 & 0.506 & 0.549 & 0.540 & 0.514 & 0.533 & 0.553 & 0.512 \\
MMAlaya2~\cite{mmalaya2} & 0.557 & 0.511 & 0.485 & 0.510 & 0.554 & 0.549 & 0.497 & 0.554 & 0.524 & 0.534 & 0.549 & 0.583 & 0.531 \\
Ovis1.5-Llama3-8B~\cite{ovis} & 0.540 & 0.505 & 0.485 & 0.499 & 0.543 & 0.526 & 0.489 & 0.530 & 0.539 & 0.518 & 0.540 & 0.565 & 0.520 \\
OmChat~\cite{omchat} & 0.459 & 0.415 & 0.394 & 0.365 & 0.455 & 0.437 & 0.432 & 0.450 & 0.457 & 0.449 & 0.449 & 0.468 & 0.432 \\
InternVL-Chat-V1-5~\cite{internvl1.5} & 0.569 & 0.522 & 0.493 & 0.518 & 0.562 & 0.556 & 0.519 & 0.556 & 0.534 & 0.546 & 0.559 & 0.581 & 0.545 \\
llava\_onevision\_qwen2\_7b\_si~\cite{llavaonevision} & 0.514 & 0.462 & 0.439 & 0.454 & 0.520 & 0.498 & 0.446 & 0.494 & 0.504 & 0.485 & 0.514 & 0.543 & 0.485 \\
XComposer2d5~\cite{xcomposer25} & 0.446 & 0.449 & 0.470 & 0.411 & 0.472 & 0.458 & 0.472 & 0.410 & 0.459 & 0.441 & 0.482 & 0.472 & 0.478 \\
Pixtral-12B~\cite{pixtral12b} & 0.567 & 0.525 & 0.513 & 0.530 & 0.574 & 0.557 & 0.518 & 0.557 & 0.554 & 0.550 & 0.561 & 0.579 & 0.547 \\
InternVL2-4B~\cite{internvl} & 0.573 & 0.524 & 0.507 & 0.534 & 0.578 & 0.564 & 0.525 & 0.540 & 0.546 & 0.540 & 0.574 & 0.587 & 0.552 \\
llava\_onevision\_qwen2\_7b\_ov~\cite{llavaonevision} & 0.504 & 0.452 & 0.439 & 0.470 & 0.512 & 0.496 & 0.432 & 0.491 & 0.491 & 0.465 & 0.513 & 0.541 & 0.478 \\
glm-4v-9b~\cite{glm2024chatglm} & 0.534 & 0.487 & 0.474 & 0.413 & 0.544 & 0.522 & 0.476 & 0.490 & 0.490 & 0.504 & 0.546 & 0.536 & 0.519 \\
molmo-7B-D-0924~\cite{molmo_pixmoopenweights} & 0.537 & 0.495 & 0.483 & 0.447 & 0.550 & 0.529 & 0.496 & 0.509 & 0.524 & 0.507 & 0.535 & 0.551 & 0.518 \\
XComposer2\_4KHD~\cite{xcomposer2_4khd} & 0.444 & 0.419 & 0.388 & 0.387 & 0.457 & 0.433 & 0.425 & 0.417 & 0.428 & 0.435 & 0.465 & 0.469 & 0.437 \\
MiniCPM-Llama3-V-2\_5~\cite{minicpm} & 0.539 & 0.498 & 0.478 & 0.485 & 0.547 & 0.534 & 0.496 & 0.523 & 0.529 & 0.526 & 0.539 & 0.555 & 0.521 \\
VILA1.5-40b~\cite{lin2023vila} & 0.500 & 0.443 & 0.417 & 0.448 & 0.495 & 0.492 & 0.447 & 0.508 & 0.458 & 0.482 & 0.506 & 0.536 & 0.497 \\
cambrian\_34b~\cite{tong2024cambrian} & 0.567 & 0.516 & 0.503 & 0.517 & 0.567 & 0.552 & 0.508 & 0.553 & 0.535 & 0.548 & 0.554 & 0.592 & 0.539 \\
WeMM~\cite{WeMM} & 0.508 & 0.465 & 0.445 & 0.437 & 0.526 & 0.493 & 0.447 & 0.496 & 0.495 & 0.467 & 0.510 & 0.519 & 0.477 \\
Llama-3.2-11B-Vision-Instruct~\cite{llama3} & 0.541 & 0.510 & 0.501 & 0.491 & 0.544 & 0.529 & 0.513 & 0.520 & 0.513 & 0.533 & 0.543 & 0.561 & 0.530 \\
Qwen2-VL-2B-Instruct~\cite{Qwen2VL} & 0.505 & 0.462 & 0.439 & 0.456 & 0.519 & 0.489 & 0.456 & 0.486 & 0.495 & 0.475 & 0.494 & 0.539 & 0.486 \\
XComposer2~\cite{xcomposer2} & 0.264 & 0.244 & 0.236 & 0.175 & 0.268 & 0.257 & 0.207 & 0.243 & 0.276 & 0.255 & 0.265 & 0.279 & 0.249 \\
cogvlm2-llama3-chat-19B~\cite{cogvlm2} & 0.486 & 0.455 & 0.445 & 0.382 & 0.506 & 0.467 & 0.440 & 0.437 & 0.458 & 0.461 & 0.496 & 0.490 & 0.464 \\
Idefics3-8B-Llama3~\cite{Idefics3} & 0.479 & 0.428 & 0.403 & 0.414 & 0.479 & 0.465 & 0.437 & 0.486 & 0.447 & 0.454 & 0.484 & 0.511 & 0.448 \\
Mini-InternVL-Chat-4B-V1-5~\cite{mini_internvl} & 0.541 & 0.496 & 0.484 & 0.488 & 0.551 & 0.533 & 0.489 & 0.534 & 0.529 & 0.522 & 0.551 & 0.562 & 0.525 \\
XinYuan-VL-2B-Instruct~\cite{XinYuan_VL_2B} & 0.393 & 0.378 & 0.365 & 0.337 & 0.427 & 0.392 & 0.387 & 0.379 & 0.395 & 0.393 & 0.416 & 0.414 & 0.398 \\
llava\_next\_yi\_34b~\cite{llavanext} & 0.498 & 0.451 & 0.432 & 0.391 & 0.511 & 0.485 & 0.429 & 0.480 & 0.475 & 0.472 & 0.489 & 0.515 & 0.461 \\
InternVL2-2B~\cite{internvl} & 0.501 & 0.450 & 0.440 & 0.395 & 0.511 & 0.479 & 0.440 & 0.469 & 0.479 & 0.453 & 0.501 & 0.498 & 0.471 \\
Phi-3-Vision~\cite{Phi_3} & 0.469 & 0.431 & 0.423 & 0.365 & 0.493 & 0.464 & 0.442 & 0.418 & 0.457 & 0.449 & 0.489 & 0.485 & 0.473 \\
cambrian\_13b~\cite{tong2024cambrian} & 0.535 & 0.474 & 0.468 & 0.465 & 0.540 & 0.508 & 0.477 & 0.502 & 0.526 & 0.521 & 0.532 & 0.546 & 0.519 \\
Phi-3.5-Vision~\cite{Phi_3} & 0.499 & 0.447 & 0.428 & 0.448 & 0.502 & 0.485 & 0.429 & 0.478 & 0.474 & 0.479 & 0.498 & 0.524 & 0.478 \\
idefics2\_8b~\cite{idefics2} & 0.467 & 0.425 & 0.408 & 0.402 & 0.488 & 0.451 & 0.402 & 0.431 & 0.443 & 0.459 & 0.474 & 0.486 & 0.451 \\
cambrian\_8b~\cite{tong2024cambrian} & 0.529 & 0.485 & 0.460 & 0.458 & 0.537 & 0.520 & 0.463 & 0.518 & 0.518 & 0.504 & 0.533 & 0.541 & 0.512 \\
minimonkey~\cite{minimonkey} & 0.467 & 0.422 & 0.413 & 0.374 & 0.488 & 0.455 & 0.441 & 0.430 & 0.459 & 0.429 & 0.478 & 0.465 & 0.454 \\
Llama-3-MixSenseV1\_1~\cite{Llama3_MixSenseV1_1} & 0.492 & 0.448 & 0.423 & 0.420 & 0.489 & 0.484 & 0.434 & 0.476 & 0.481 & 0.466 & 0.490 & 0.526 & 0.459 \\
llava\_next\_qwen\_32b~\cite{llavanext} & 0.526 & 0.479 & 0.456 & 0.449 & 0.539 & 0.516 & 0.468 & 0.494 & 0.514 & 0.489 & 0.528 & 0.543 & 0.501 \\
xgen-mm-phi3-interleave-r-v1.5~\cite{xgenmm} & 0.549 & 0.503 & 0.492 & 0.495 & 0.550 & 0.533 & 0.515 & 0.554 & 0.529 & 0.511 & 0.549 & 0.562 & 0.524 \\
Eagle-X5-13B~\cite{Eagle} & 0.446 & 0.420 & 0.390 & 0.387 & 0.455 & 0.440 & 0.407 & 0.433 & 0.424 & 0.440 & 0.458 & 0.443 & 0.439 \\
xgen-mm-phi3-dpo-r-v1.5~\cite{xgenmm} & 0.420 & 0.399 & 0.391 & 0.289 & 0.464 & 0.414 & 0.402 & 0.409 & 0.412 & 0.430 & 0.468 & 0.438 & 0.415 \\
Mini-InternVL-Chat-2B-V1-5~\cite{mini_internvl} & 0.447 & 0.408 & 0.394 & 0.308 & 0.474 & 0.433 & 0.412 & 0.410 & 0.430 & 0.416 & 0.455 & 0.439 & 0.426 \\
llava\_next\_llama3~\cite{llavanext} & 0.495 & 0.455 & 0.427 & 0.426 & 0.514 & 0.485 & 0.441 & 0.473 & 0.478 & 0.474 & 0.493 & 0.516 & 0.474 \\
VILA1.5-13b~\cite{lin2023vila} & 0.499 & 0.458 & 0.432 & 0.473 & 0.496 & 0.491 & 0.468 & 0.517 & 0.468 & 0.484 & 0.502 & 0.508 & 0.486 \\
Mantis-8B-siglip-llama3~\cite{Mantis} & 0.424 & 0.408 & 0.376 & 0.365 & 0.418 & 0.409 & 0.411 & 0.406 & 0.401 & 0.433 & 0.441 & 0.421 & 0.422 \\

\bottomrule
\end{tabular}
}
\caption{Per-pattern SIM on MMGenBench-Test for all 68 LMMs. The icons are explained in Fig.~\ref{fig:dataset-case}. Patterns overlap; values are descriptive means. Model references and overall results appear in Table~\ref{tab:suppl-results-all}.}
\label{tab:suppl-results-test}
\end{table*}\space{}
\begin{table*}[th]
\centering
\begin{tabular}{lcccccc}
\toprule
LMM & Mean & Std & Min & Max & Range & \#Models \\
\midrule
Eagle-X5-13B            & 0.4484 & 0.0263 & 0.4249 & 0.4849 & 0.0599 & 4 \\
XComposer2              & 0.2617 & 0.0260 & 0.2348 & 0.2947 & 0.0599 & 4 \\
XinYuan-VL-2B-Instruct  & 0.4060 & 0.0231 & 0.3832 & 0.4371 & 0.0539 & 4 \\
idefics2\_8b            & 0.4567 & 0.0210 & 0.4354 & 0.4852 & 0.0498 & 4 \\
Mantis-8B-siglip-llama3 & 0.4189 & 0.0206 & 0.3974 & 0.4462 & 0.0488 & 4 \\
Phi-3-Vision            & 0.4657 & 0.0198 & 0.4461 & 0.4928 & 0.0467 & 4 \\
minimonkey              & 0.4579 & 0.0192 & 0.4384 & 0.4838 & 0.0454 & 4 \\
VILA1.5-40B             & 0.4874 & 0.0181 & 0.4715 & 0.5134 & 0.0419 & 4 \\
OmChat                  & 0.4453 & 0.0172 & 0.4299 & 0.4698 & 0.0399 & 4 \\
xgen-mm-phi3-dpo-r-v1.5 & 0.4165 & 0.0163 & 0.3968 & 0.4366 & 0.0398 & 4 \\
\bottomrule
\end{tabular}
\caption{LMMs with the largest cross-generator SIM standard deviations.}
\label{sensitivity_analysis_of_t2i_model:6}
\end{table*}

\begin{table}[ht]
\centering
\resizebox{\columnwidth}{!}{\begin{tabular}{lcccc}
\toprule
            & Flux-1-dev & SD3.5 & Kolors & Lumina \\
\midrule
Flux-1-dev & 1.0000 & 0.9912 & 0.9769 & 0.9851 \\
SD3.5      & 0.9912 & 1.0000 & 0.9679 & 0.9779 \\
Kolors     & 0.9769 & 0.9679 & 1.0000 & 0.9844 \\
Lumina     & 0.9851 & 0.9779 & 0.9844 & 1.0000 \\
\bottomrule
\end{tabular}
}
\caption{Spearman correlations of LMM rankings across T2I models based on average SIM-Score.}
\label{sensitivity_analysis_of_t2i_model:1}
\end{table}

\begin{table}[ht]
\centering
\resizebox{\columnwidth}{!}{\begin{tabular}{lcccc}
\toprule
            & Flux-1-dev & SD3.5 & Kolors & Lumina \\
\midrule
Flux-1-dev & 1.0000 & 0.9394 & 0.8828 & 0.9111 \\
SD3.5      & 0.9394 & 1.0000 & 0.8626 & 0.8909 \\
Kolors     & 0.8828 & 0.8626 & 1.0000 & 0.9152 \\
Lumina     & 0.9111 & 0.8909 & 0.9152 & 1.0000 \\
\bottomrule
\end{tabular}
}
\caption{Kendall correlations of LMM rankings across T2I models based on average SIM-Score.}
\label{sensitivity_analysis_of_t2i_model:2}
\end{table}

\begin{table}[ht]
\centering
\resizebox{\columnwidth}{!}{\begin{tabular}{lcccc}
\toprule
            & Flux-1-dev & SD3.5 & Kolors & Lumina \\
\midrule
Flux-1-dev & 1.0000 & 0.9679 & 0.8631 & 0.9799 \\
SD3.5      & 0.9679 & 1.0000 & 0.7956 & 0.9455 \\
Kolors     & 0.8631 & 0.7956 & 1.0000 & 0.9013 \\
Lumina     & 0.9799 & 0.9455 & 0.9013 & 1.0000 \\
\bottomrule
\end{tabular}
}
\caption{Spearman correlations of LMM rankings across T2I models based on FID-Score. FID estimates on MMGenBench-Test and its patterns are exploratory.}
\label{sensitivity_analysis_of_t2i_model:3}
\end{table}

\begin{table}[ht]
\centering
\resizebox{\columnwidth}{!}{\begin{tabular}{lcccc}
\toprule
            & Flux-1-dev & SD3.5 & Kolors & Lumina \\
\midrule
Flux-1-dev & 1.0000 & 0.8734 & 0.7131 & 0.8882 \\
SD3.5      & 0.8734 & 1.0000 & 0.6458 & 0.8182 \\
Kolors     & 0.7131 & 0.6458 & 1.0000 & 0.7495 \\
Lumina     & 0.8882 & 0.8182 & 0.7495 & 1.0000 \\
\bottomrule
\end{tabular}
}
\caption{Kendall correlations of LMM rankings across T2I models based on FID-Score. FID estimates on MMGenBench-Test and its patterns are exploratory.}
\label{sensitivity_analysis_of_t2i_model:4}
\end{table}

\begin{table}[ht]
\centering
\begin{tabular}{lcc}
\toprule
Metric & Method & Avg. Correlation \\
\midrule
SIM & Spearman & 0.9806 \\
SIM & Kendall  & 0.9003 \\
FID & Spearman & 0.9089 \\
FID & Kendall  & 0.7814 \\
\bottomrule
\end{tabular}
\caption{Average correlation coefficients of LMM rankings across T2I models.}
\label{sensitivity_analysis_of_t2i_model:5}
\end{table}

\begin{table}[ht]
\centering
\begin{tabular}{lcc}
\toprule
Metric & Statistic & Value \\
\midrule
SIM & Average Range & 0.0301 \\
SIM & Average Std   & 0.0132 \\
FID & Average Range & 0.3138 \\
FID & Average Std   & 0.1389 \\
\bottomrule
\end{tabular}
\caption{Mean cross-generator standard deviations and ranges of LMM scores.}
\label{sensitivity_analysis_of_t2i_model:7}
\end{table}\space{}
\begin{table}[!ht]
\centering
\begin{tabular}{lccc}
\toprule
        & clip   & dino   & unicom \\
\midrule
clip    & 1.0000 & 0.9788 & 0.9835 \\
dino    & 0.9788 & 1.0000 & 0.9851 \\
unicom  & 0.9835 & 0.9851 & 1.0000 \\
\bottomrule
\end{tabular}
\caption{SIM-Score: Spearman correlations of LMM rankings across embedding models.}
\label{sensitivity_analysis_of_embedding_model:1}
\end{table}

\begin{table}[!ht]
\centering
\begin{tabular}{lccc}
\toprule
        & clip   & dino   & unicom \\
\midrule
clip    & 1.0000 & 0.8869 & 0.8990 \\
dino    & 0.8869 & 1.0000 & 0.9098 \\
unicom  & 0.8990 & 0.9098 & 1.0000 \\
\bottomrule
\end{tabular}
\caption{SIM-Score: Kendall correlations of LMM rankings across embedding models.}
\label{sensitivity_analysis_of_embedding_model:2}
\end{table}

\begin{table}[!ht]
\centering
\begin{tabular}{lccc}
\toprule
        & clip   & dino   & unicom \\
\midrule
clip    & 1.0000 & 0.9891 & 0.9898 \\
dino    & 0.9891 & 1.0000 & 0.9876 \\
unicom  & 0.9898 & 0.9876 & 1.0000 \\
\bottomrule
\end{tabular}
\caption{FID-Score: Spearman correlations of LMM rankings across embedding models. FID estimates on MMGenBench-Test and its patterns are exploratory.}
\label{sensitivity_analysis_of_embedding_model:3}
\end{table}

\begin{table}[!ht]
\centering
\begin{tabular}{lccc}
\toprule
        & clip   & dino   & unicom \\
\midrule
clip    & 1.0000 & 0.9246 & 0.9273 \\
dino    & 0.9246 & 1.0000 & 0.9138 \\
unicom  & 0.9273 & 0.9138 & 1.0000 \\
\bottomrule
\end{tabular}
\caption{FID-Score: Kendall correlations of LMM rankings across embedding models. FID estimates on MMGenBench-Test and its patterns are exploratory.}
\label{sensitivity_analysis_of_embedding_model:4}
\end{table}

\space{}
\clearpage

\space{}
\begin{figure*}[ht]
\centering
\begin{tcolorbox}[title=Extracted Image Patterns]
\small
    ``Surreal": 2262,
    ``Structural and Physical Characteristics": 2016,
    ``Position and Contextual Relationship": 1775,
    ``Geometry": 1270,
    ``Orientation and Direction": 1260,
    ``State and Condition": 1119,
    ``Text": 907,
    ``Color": 420,
    ``Color Contrast": 404,
    ``Quantity": 355,
    ``Color Scheme": 172,
    ``Natural Elements": 152,
    ``Nature": 148,
    ``Contextual Relationship": 137,
    ``Symmetry": 126,
    ``Texture": 117,
    ``Anthropomorphism": 103,
    ``Light and Shadow": 96,
    ``Fantasy": 86,
    ``Color Gradient": 81,
    ``Symbolism": 80,
    ``Color and Light": 72,
    ``Lighting": 70,
    ``Technology": 67,
    ``Geometric Shapes": 64,
    ``Contrast": 64,
    ``Architecture": 62,
    ``Ornamentation": 59,
    ``Nature Elements": 59,
    ``Light and Illumination": 55,
    ``Lighting and Illumination": 54,
    ``Color and Texture": 54,
    ``Reflection": 49,
    ``Color Pattern": 49,
    ``Clothing and Accessories": 46,
    ``Color Palette": 44,
    ``Color and Contrast": 40,
    ``Lighting and Color": 37,
    ``Light and Color": 37,
    ``Textural Elements": 35,
    ``Textural": 34,
    ``Geometric": 34,
    ``Color Patterns": 32,
    ``Animal Representation": 32,
    ``Floral": 32,
    ``Historical Context": 31,
    ``Abstract": 31,
    ``Lighting and Shadow": 30,
    ``Repetition": 29,
    ``Fashion": 29,
    ``Accessories": 28,
    ``Cultural Elements": 28,
    ``Geometric Patterns": 26,
    ``Color and Lighting": 26,
    ``Character Design": 25,
    ``Decorative Elements": 25,
    ``Textural Details": 25,
    ``Textural and Physical Characteristics": 24,
    ``Ornamental Design": 23,
\\
......
\\
    ``Medieval and Fantasy Aesthetics": 1,
    ``Symbolism and Gestures": 1,
    ``Mechanics and Cybernetics": 1,
    ``Imagery and Graphics": 1,
    ``Detail and Complexity": 1,
    ``Cultural Representations": 1,
    ``Fusion of Elements": 1,
    ``Cartoon-like Aesthetic": 1
\end{tcolorbox}

\caption{Image patterns extracted by GPT-4o and ranked in descending order of frequency.}
\label{tc:extracted-image-patterns}
\end{figure*}\space{}
\begin{figure*}[ht]
\centering
\begin{tcolorbox}[title=Prompt for Extraction]
\# Role
\par
You are an expert in annotating patterns of images, specializing in the analysis and annotation of images.
\par\medskip
\# Definition Explanation
\par
Image Patterns: Refer to repetitive or consistent visual features or structures that appear in images. These patterns can include colors, shapes, textures, arrangements, or even specific objects or symbols.
\par
Example Patterns: "Orientation and Direction", "Text", "Quantity", "Geometry", "Surreal", "State and Condition", "Position and Contextual Relationship", etc.
\par\medskip
\# Task Instructions
\par
1. Generate a detailed description of the image based on its content, including all details observed in the image.
\par
2. Annotate possible patterns in the image based on the image elements and the description of the image, and explain the reasoning for each identified pattern.
\par\medskip
\# Key Points
\par
Carefully inspect all details within the image and annotate possible patterns for the image.
\par
Patterns should be based on the contents and visual elements of the image. You may annotate multiple patterns as appropriate.
\par\medskip
\# Output Format
\par
A JSON composed of descriptions, patterns, and explanations. The elements are divided into strings, lists of strings, and dictionaries. For example:
\par
\{
    "description": "This is a surreal black and white painting depicting a person holding a pile of brains with both hands. The shapes and textures of the brains are clearly visible and exhibit a merging effect. The entire image carries a mysterious and subtly eerie feeling.",
    ``image\_pattern'': [``Text'', ``Structural and Physical Characteristics''],
    ``pattern\_detail'': \{
        ``Text'': ``The image contains the text `dog'.'',
        ``Structural and Physical Characteristics": "The image includes buildings with prominent leaning features.''
\}\}
\end{tcolorbox}
\caption{Prompt for image pattern extraction. The JSON example illustrates the output structure; it is not a validated image annotation.}
\label{tc:Prompt-for-Extraction}
\end{figure*}\space{}
\begin{figure*}[ht]
\centering
\begin{tcolorbox}[title=Prompt for Summary]
\# Role
\par
You are an expert in image pattern analysis, focusing on the extraction and summarization of image patterns.
\par\medskip
\# Definition Explanation
\par
Image Patterns: Refer to repetitive or consistent visual features or structures that appear in images. These patterns can include colors, shapes, textures, arrangements, or even specific objects or symbols.
\par
Example Patterns: ``Text'', ``Quantity'', ``Geometry'', ``Surreal'', etc.
\par\medskip
\# Task Instructions
\par
1. Summarize the optimal pattern list based on the input JSON data (including image patterns and frequencies).
\par
2. Describe the explanations for summarizing the pattern list.
\par\medskip
\# Key Points
\par
Carefully examine the image patterns and their frequencies in the input data, summarize a new list of image patterns, and provide reasons.
\par
The new image pattern words should be as short as possible.
\par
Not only should you rely on frequency, but also abstract summarization, to abstract from the existing image patterns.
\par\medskip
\# Input Format
\par
JSON data, the key is the image pattern, and the value is the frequency the image pattern appears. Example:
\par
\{
    ``Surreal'': 2262,
    ``Lighting'': 70,
    ``Lighting and Illumination'': 54,
\}
\par
The input data indicates that the image pattern `Surreal` appears 2262 times, the image pattern `Lighting` appears 70 times, and the image pattern `Lighting and Illumination` appears 54 times.
\par\medskip
\# Output Format
\par
A JSON composed of patterns and explanations. The elements are divided into lists of strings and dictionaries. Example:
\par
\{
  ``image\_pattern'': [``Surreal'', ``Lighting''],
  ``pattern\_detail'': \{
    ``Surreal'': ``Surreal appears frequently'',
    ``Lighting'': ``Lighting appears multiple times, and is synonymous with Illumination.''
\}\}
\end{tcolorbox}
\caption{Prompt for image pattern summarization.}
\label{tc:Prompt-for-Summary}
\end{figure*}\space{}
\begin{figure*}[ht]
\centering
\begin{tcolorbox}[title=Prompt for Re-annotation]
\small
\# Role
\par
You are an expert in annotating patterns of images, specializing in the analysis and annotation of images.
\par\smallskip
\# Definition Explanation
\par
Image Patterns: Refer to repetitive or consistent visual features or structures that appear in images. These patterns can include colors, shapes, textures, arrangements, or even specific objects or symbols.
\par
Example Patterns: ``Surreal'', ``Text'', ``Count'', etc.
\par\smallskip
\# Task Instructions
\par
1. Generate a detailed description of the image based on its content, including all details observed in the image.
\par
2. Annotate possible patterns in the image based on the image elements and the description of the image, and explain the reasoning for each identified pattern.
\par\smallskip
\# Key Points
\par
1. Carefully inspect all details within the image and annotate possible patterns for the image.
\par
2. Patterns should be based on the contents and visual elements of the image. You may annotate multiple patterns as appropriate.
\par
3. The labeling pattern can only come from the following patterns:
\par
\{
    ``Surreal'': ``This pattern is characterized by its prevalence in depicting scenes that mix elements of fantasy with reality, often creating imaginative or dream-like visuals.'',
    ``Technology'': ``Highlights themes related to technology, futurism, and modernity, encompassing various aspects of technological advances and speculative futures.'',
    ``Natural'': ``Encompasses imagery related to natural sceneries or elements, drawing on the beauty and complexity of the natural world.'',
    ``Artistic'': ``Captures various artistic styles and decorations, reflecting creativity, design elements, and unique art styles.'',
    ``Color'': ``Focuses on the use and significance of color in images, including aspects like color contrast, schemes, and gradients to convey moods or themes.'',
    ``Count'': ``Deals with patterns that emphasize numerical aspects, quantities, and the presence of multiple elements, indicating a focus on enumeration or amount.'',
    ``Orientation'': ``Emphasizes the directionality and alignment within images, capturing how subjects or elements are oriented or directed.'',
    ``Position'': ``Concerns the relative placement and spatial relationships between elements within an image, highlighting how objects are positioned.'',
    ``Contextual'': ``Fuses themes of environmental and relational context, shedding light on the settings and the interconnections within images.'',
    ``Text'': ``Identifies patterns where text is a significant component, showcasing how textual content contributes to the overall message or theme of the image.'',
    ``Symbol'': ``Centers on the use of symbols and symbolic elements to convey deeper meanings or concepts, drawing from various symbolic traditions.'',
    ``Geometry'': ``Illustrates a focus on geometric shapes, patterns, and arrangements, underlining the role of structure and form in images.'',
    ``Motion'': ``Captures the sense of movement or dynamics within images, depicting actions or the concept of motion through visual elements.''
\}
\par
Where key is the image pattern and value is the explanation of the image pattern.
\par\smallskip
\# Output Format
\par
A JSON composed of descriptions, patterns, and explanations. The elements are divided into strings, lists of strings, and dictionaries. For example:
\par
\{
    ``description'': ``This is a surreal black and white painting depicting a person holding a pile of brains with both hands. The shapes and textures of the brains are clearly visible and exhibit a merging effect. The entire image carries a mysterious and subtly eerie feeling.'',
    ``image\_pattern'': [``Text'', ``Count''],
    ``pattern\_detail'': \{
      ``Text'': ``The image contains the text `dog'.'',
      ``Count'': ``The image contains 3 dogs.''
\}\}
\end{tcolorbox}
\caption{Prompt for image pattern re-annotation. The JSON example illustrates the output structure; it is not a validated image annotation.}
\label{tc:Prompt-for-Re-annotation}
\end{figure*}\space{}
\begin{figure*}[ht]
\centering
\begin{tcolorbox}[title=Evaluation Pipeline Prompt]
\# Role
\par
You are an expert in the field of image understanding, focusing on the understanding of images and generating the image caption-prompt.
\par\medskip
\# Definition Explanation
\par
image caption-prompt: Refers to the caption or description of an image, used to provide to a Text-to-Image model to generate a new image.
\par
Text-to-Image model: Can generate a new image based on the provided image caption-prompt, such as stable diffusion 3, flux, and other image generation models.
\par\medskip
\# Task Description
\par
Generate an image caption-prompt based on the input image.
\par\medskip
\# Key Points and Requirements
\par
1. Accurately understand the input image and precisely generate an image caption-prompt.
\par
2. The generated image caption-prompt, when provided to the Text-to-Image model, requires the Text-to-Image model to generate a new image that is as consistent as possible with the input image.
\par
3. The generated image caption-prompt must conform to the preferences of the Text-to-Image model.
\par
4. The generated image caption-prompt should describe the input image in as much detail as possible, and it should be between 20 to 60 words.
\par\medskip
\# Output Format
\par
A string, that is the image caption-prompt. No extra output needed.
\end{tcolorbox}
\caption{Evaluation pipeline prompt corresponding to the multimodal input in Fig.~\ref{fig:pipeline}.}
\label{tc:Evaluation-Pipeline-Prompt}
\end{figure*}\space{}
\space{}
\space{}
\space{}
\space{}
\space{}\space{}
\end{document}